%% file: neurips_2025.tex
\documentclass{article}

\usepackage[preprint]{neurips_2025}

\PassOptionsToPackage{numbers}{natbib}
\setcitestyle{square,numbers,comma}

\usepackage[utf8]{inputenc} % allow utf-8 input
\usepackage[T1]{fontenc}    % use 8-bit T1 fonts
\usepackage{hyperref}       % hyperlinks
\usepackage{url}            % simple URL typesetting
\usepackage{booktabs}       % professional-quality tables
\usepackage{amsfonts}       % blackboard math symbols
\usepackage{nicefrac}       % compact symbols for 1/2, etc.
\usepackage{microtype}      % microtypography
\usepackage{xcolor}         % colors

\usepackage{graphicx}
\usepackage{subfigure}
\usepackage{booktabs} 
\usepackage{multirow}
\usepackage{mathtools}
\usepackage{amsthm}
\usepackage{amsmath}
\usepackage{amssymb}
\usepackage{wrapfig}
\usepackage[ruled,vlined,linesnumbered]{algorithm2e}

\newcommand\boldsection[1]{\vspace{6pt}{\noindent\bf #1.}}

\DeclareRobustCommand{\MD}{\textbf{\textit{MotifDisco}}\xspace}

\newcommand{\Autoref}[1]{%
  \begingroup%
  \def\chapterautorefname{Chapter}%
  \def\sectionautorefname{Section}%
  \def\subsectionautorefname{Section}%
  \def\algorithmautorefname{Algorithm}%
  \autoref{#1}%
  \endgroup%
}

\title{\MD: Motif Causal Discovery For Time Series Motifs}

\author{%
  Josephine Lamp, Mark Derdzinski, Christopher Hannemann, \\\textbf{Sam Hatfield, Joost van der Linden} \\
  Dexcom, USA\thanks{Disclaimer: The views and opinions expressed in this paper are those of the authors and do not necessarily reflect the official policy or position of Dexcom Inc.}
}

\begin{document}

\maketitle

\begin{abstract}
Many time series, particularly health data streams, can be best understood as a sequence of phenomenon or events, which we call \textit{motifs}. A time series motif is a short trace segment which may implicitly capture an underlying phenomenon within the time series. Specifically, we focus on glucose traces collected from continuous glucose monitors (CGMs), which inherently contain motifs representing underlying human behaviors such as eating and exercise. The ability to identify and quantify \textit{causal} relationships amongst motifs can provide a mechanism to better understand and represent these patterns, useful for improving deep learning and generative models and for advanced technology development (e.g., personalized coaching and artificial insulin delivery systems). However, no previous work has developed causal discovery methods for time series motifs. Therefore, in this paper we develop \MD (\textbf{motif} \textbf{disco}very of causality), a novel causal discovery framework to learn causal relations amongst motifs from time series traces. We formalize a notion of \textit{Motif Causality (MC)}, inspired from Granger Causality and Transfer Entropy, and develop a Graph Neural Network-based framework that learns causality between motifs by solving an unsupervised link prediction problem. We integrate MC with three model use cases of forecasting, anomaly detection and clustering, to showcase the use of MC as a building block for downstream tasks. Finally, we evaluate our framework on different health data streams and find that Motif Causality provides a significant performance improvement in all use cases.
\end{abstract}

%9 pages of full content

\input{1.intro}
\input{2.related}
\input{3.approach}

\input{4.eval}
\input{5.conclusion}

\bibliography{references}

%%%%%%%%%%%%%%%%%%%%%%%%%%%%%%%%%%%%%%%%%%%%%%%%%%%%%%%%%%%%

\input{6.appendix}

\end{document}

%% file: 1.intro.tex
\section{Introduction}\label{sec:intro}

% % Many time series are event-guided, meaning they 
% Many time series can be best understood as a sequence of phenomenon or events, which we call \textit{motifs}. A motif is a short chunk of a trace 
% % short, ordered sequence of values in a time series 
% which may implicitly capture an underlying behavior or phenomenon within the time series. 
% This extremely common and relevant in many health data streams, where traces are guided by underlying human physiology or behaviors. Examples include hormone traces, activity streams, inpatient event sequences, and hormone traces. 

%% MOTIFS
Many time series can be best understood as sequences of phenomenon or events. This is extremely common in many health data streams where traces are guided by underlying human physiology or behaviors. 
% Examples include hormone traces, activity streams, inpatient event sequences, and hormone traces. 
We call these events in the traces \textit{motifs}. A time series motif is a short trace segment 
% short, ordered sequence of values in a time series 
which may implicitly capture an underlying behavior or phenomenon within the time series. 
To contextualize our discussion, and as our main running example, we focus on glucose traces collected from continuous glucose monitors (CGMs) for diabetes. Glucose traces inherently contain motifs which represent underlying human behaviors. For instance, a motif capturing a peak in glucose may correspond to an individual eating; a motif capturing a drop in glucose may correspond to an individual exercising. 
\Autoref{fig:sample-motifs} shows real glucose traces and motifs  
(other examples in Appx.~\ref{sec:appdx-intro}).
% also applicable to other areas such as weather streams, logging streams (such as in networks or intrusion detection systems), 
% Glucose traces can be best understood as sequences of events, so can many other time series and particularly health data streams.

%% Causal relations in motifs helpful for stuff
Causal discovery is the process by which causal relations are found amongst observational data \citep{niu2024comprehensive}. 
The ability to discover and quantify 
causality amongst motifs 
% causal relationships amongst motifs 
can provide a mechanism to better understand and represent these patterns, useful in a variety of applications.
For instance, learning causal relations in glucose motifs may enable better understanding of physiological 
% and behavioral
patterns contributing to advanced technology development (e.g., artificial insulin delivery systems). 
Moreover, motif causal relationships may be helpful building blocks when used as sub-components in generative and deep learning models to improve their performance for other downstream tasks.
% and model architectures (e.g., to improve deep learning and generative models),
% to help improve the accuracy of generative and deep learning models used for other downstream tasks.
% , such as classification, forecasting, generative models, synthetic data use cases, and many other generative and deep learning models trying to predict or represent these types of traces.

% In other words, use of motif causality can be used as a building block for other model architectures. This idea is similar to how 
%     Autoencoders - important to represent and embed complex data
%     Autoregressive - important to understand temporal relationships amongst sequences
%     Motif Caus - equivalent to this for event based seqs

%% Other work in Causal Discovery
% Causal graphs are generated/discovered from data enabling the identification and estimation of causal effects
% There has been a lot of work for multivariate time series causal discovery \citep{gong2023causal, assaad2022survey}. 
%Motif causality is a new notion which helps to better capture underlying mechanics and behaviors of event based time series which could be helpful as a building block for other models

Granger Causality \citep{granger1969investigating} and its nonlinear extension Transfer Entropy (TE) \citep{schreiber2000measuring} are commonly used in time series causal discovery methods, as causal relations are quantified based on one trace's \textit{predictability} of another. This notion of causality provides an intuitive, understandable measure to quantify 
% and understand 
causal relations 
% grounded in information theoretic measures 
and is advantageous for time series because it innately incorporates temporality without requiring strong model assumptions.
% does not require strong model assumptions, and innately incorporates notions of temporality.
% Many methods use Granger Causality \citep{granger1969investigating} or its nonlinear extension Transfer Entropy (TE) \citep{schreiber2000measuring}, which quantifies causal relations based on \textit{predictability}. 
% Transfer Entropy (TE) is a nonlinear extension of Granger Causality and quantifies causality using information theoretic measures in terms of uncertainty reduction \citep{shojaie2022granger}. 
% A popular causal framework  Granger causality~\cite{granger1969investigating}, a useful causal notion for time series because causality is defined in terms of predictability. 
Despite exciting recent developments for time series causal discovery \citep{gong2023causal, assaad2022survey}, no previous work has focused on causal discovery amongst time series \textit{motifs}.
% \textit{motifs} contained within time series traces. 
% However, there have been no works developed specifically for event-based time series or motifs, shapelets, anything of the type.
Motif causal discovery is challenging because 
in many cases (particularly in health) 
there is no ground truth about what underlying behavior a motif captures.
% may be captured by a specific motif itself.
For example, 
% in diabetes underlying glucose changes are not labelled by behaviors (a rise in glucose may be from eating or it may be from stress.) 
from data alone one cannot conclusively determine what caused a change in glucose
% one cannot conclusively determine what behavior caused each change in glucose from glucose data alone 
(e.g., a glucose rise may be from eating \textit{or} stress).
% (e.g., a rise in glucose may be from eating \textit{or} it may be from stress).
% Moreover, it would be impossible to label every change in glucose (as this would require an individual to document every single behavior they conducted throughout the day, which is not realistic, and in some cases one may not even know why a glucose change occurred due to innate human physiology). 
%In addition,
As a result, unlike in many other causal models, there is no ground truth causal structure amongst motifs that could be used to guide the casual discovery model 
% ground truth causal structure amongst motifs is not known and cannot be used to guide causal discovery model training 
(i.e., through supervised methods using labeled causal events.) 
% so there is no way to conclusively validate the learned causal graphs.
% Irregularness of them, lack of temporality within time series in some cases

Therefore, in this paper we develop \MD, (\textbf{motif} \textbf{disco}very of causality), a framework to discover causal relations from time series motifs.
% amongst a set of ordered motifs pulled from time series traces. 
First, we formalize the concept of motifs and define a notion of \textit{Motif Causality} (MC) inspired from Granger Causality and Transfer Entropy, which is able to characterize causal relationships amongst sequences of motifs.
Next, we develop a causal discovery framework to learn MC amongst a set of ordered motifs pulled from time series traces. The framework uses a Graph Neural Network (GNN) based architecture that learns causality amongst motifs by solving an unsupervised link prediction problem, thereby not requiring knowledge of any ground truth causal structure for training.
The framework outputs a directed causal graph where nodes represent motifs and edges represent the degree of the MC relationship.
% uses an unsupervised training of a Graph Neural Network (GNN) architecture incorporated with Motif Causality computation to learn and output a causal graph where nodes represent motifs and edges represent the degree of motif-causal relation between the nodes. 
To demonstrate the suitability of Motif Causality as a building block in other models for downstream tasks, 
we instantiate three model use cases that incorporate 
% the learned motif causal graphs 
MC
for forecasting, anomaly detection, and clustering tasks.
% We use motif causality / learned motif causal graphs in three application areas, including forecasting, anomaly detection and clustering, to showcase the suitability of motif causality as a building block for other models / downstream purposes.
Finally, we evaluate \MD in terms of scalability and use case performance by comparing the models with and without integration of MC, to see how helpful MC is for each use case.

The contributions of this paper are: (1) We formalize a new notion of causality between time series motifs, denoted as Motif Causality. 
% notion of Motif Causality inspired from Granger Causality and Transfer Entropy. 
(2) We develop \MD, the first causal discovery framework to learn Motif Causality amongst time series motifs.  
% a novel GNN-based framework to learn motif causality from a set of input time series traces. 
(3) We illustrate the use of MC as a building block in other downstream tasks by integrating MC with three model use cases of forecasting, anomaly detection and clustering.
% instantiate three model use cases that integrate with motif causality in model training to showcase the use of motif causality as a building block for other downstream tasks.
(4) We provide detailed framework evaluation 
% in terms of scalability and use case performance 
and find that MC provides a significant performance improvement compared to the base models in all use cases.

%% file: 2.related.tex
\section{Related Work}\label{sec:rel-work}

\begin{figure}[t]
    \centering
    \includegraphics[width=\linewidth]{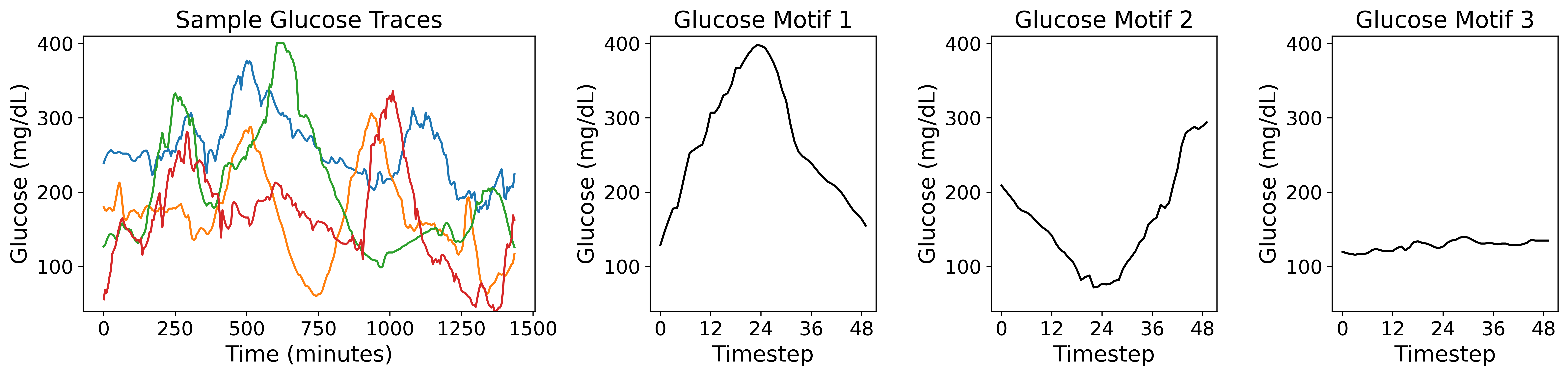}
    \caption{Real Glucose Traces and Sample Motifs for $\tau=48$.}
    \label{fig:sample-motifs}
\end{figure}

\boldsection{Motifs}
% network motifs for various causality based tasks ranging from causal discovery to use in stuff like pharmacovig, anaomaly detection etc.
Recently, there has been interest in temporal \textit{network motifs}, which are sets of recurring graph substructures. 
% There is a large body of work surrounding \textit{network motifs}, which is a different definition of motif than ours. A network motif or temporal network motif is a set of recurring graph substructures, as opposed to a behavior within the time series data itself, as is defined in our work.
Previous work has investigated network motif causality \citep{liu2021temporal, kovanen2011temporal} and  
developed network motif causal discovery 
% and graph representation learning 
frameworks \citep{chen2024tempme, chen2023motif, jin2022neural}.
% and developed 
% % new use case architectures 
% methods that utilize network motif causal structures, such as for pharmacovigilance and anomaly detection tasks \citep{xiao2024motif, kalla2023exploring, yu2022motifexplainer, zhu2022higher, mu2021disentangled}.
% such as for finding drug-drug interactions in pharmacovigilance use cases \citep{kalla2023exploring} and anomaly detection \citep{xiao2024motif}.
Importantly, the definition of motif used here is different than ours, referring to patterns in graph structures as opposed to patterns in the traces themselves.
% In other motif application areas,  \citet{chinpattanakarn2024framework} solve a different problem and develop a method that, given two time series, infers a set of patterns, also called motifs, that follow each other in the traces. 
In other motif application areas,  \citet{chinpattanakarn2024framework} solve a different problem and develop a method to infer a set of patterns, also called motifs, that follow each other in the traces. 
% i.e., occur in both time series within an arbitrary time delay). 
% Their framework uses matrix profiling to retrieve the following-relation motifs.
Finally, \citet{lamp2024glucosynth} develop a method to generate synthetic time series glucose traces 
% using Generative Adversarial Networks (GANs) 
and use a notion of causality amongst motifs to help the model perform well. 
The focus of this work is not on causality, 
% and no additional analysis or evaluation of the causality component is studied.
% Moreover, their method to 
and the causal learning method is 
% and the method used to learn causality is 
complicated and suffers from  scalability issues. 

% GlucoSynth issues
% - not explainable / hard to understand (what does motif causality actually mean ??)
% - Not scalable, takes forever to train
% - motif caus only used as component for overall synth data generation, focus not on it

% There is a large body of work surrounding \textit{network motifs}, which is a different definition of motif than ours. A network motif or temporal network motif is a set of recurring graph substructures, as opposed to a behavior within the time series data itself, as is defined in our work.
% Importantly, their definition of a motif is different than ours, and is defined as a set of recurring substructures within a graph, as opposed to a behavior in the time series data itself.

\boldsection{Causal Discovery in Time Series} 
%these surveys
There are a variety of works on causal discovery for time series~\citep{niu2024comprehensive, gong2023causal, hasan2023survey, assaad2022survey, shojaie2022granger}.
%hasan one meh can remove if need space
In particular, previous methods have developed causal discovery frameworks for multivariate time series that incorporate temporal dynamics and use Granger Causality \citep{pan2024effcause, tank2021neural, lowe2022amortized}
or Transfer Entropy \citep{bonetti2024causal, najafi2023entropy}.
Recent work has also incorporated time series causality measures to improve model learning in downstream tasks such as forecasting and anomaly detection \citep{ansari2024chronos, chen2023multi, febrinanto2023entropy, duan2022multivariate, wu2021event2graph}. 
Previous methods for time series causal discovery cannot be directly applied to motifs because they formulate causality using multivariate statistical properties 
(e.g., variable-based correlation) 
or temporal statistical measures repeated across time series lags, which do not hold for short, univariate time series motifs that do not contain repeated lagged patterns; or use labels or known underlying causal structures, which are not available for our traces and many similar event-based data streams. 
% , which do not hold for univariate time series motifs; (b) 
% temporal statistical measures repeated across time series lags 
% % (e.g., learn implicit repeated patterns or sufficiency hypotheses over multiple lags in a trace) 
% that do not work for short time series motifs which do not contain repeated patterns; or (c) labels or a known underlying causal structure to guide model training, which is not available for our traces and many similar event-based data streams.
% However, no previous work has developed causal discovery frameworks for time series motifs.
% Few previous work focused on specified "event" based time series and none focus on causality for something like motifs.
\MD is the first framework for causal discovery in time series motifs, 
% which may broaden capabilities and improve performance of current models for applications that use event-based time series.
% motif-based traces.
which may broaden current model capabilities 
% and performance for applications that use 
for event-based time series.

%% file: 3.approach.tex
\section{Formalizing Motif Causality}\label{sec:define-motif-caus}

\begin{figure*}
     \centering
      \subfigure[Chopping]{
         \centering
         \label{fig:motif-chop}
         \includegraphics[width=0.31\textwidth]{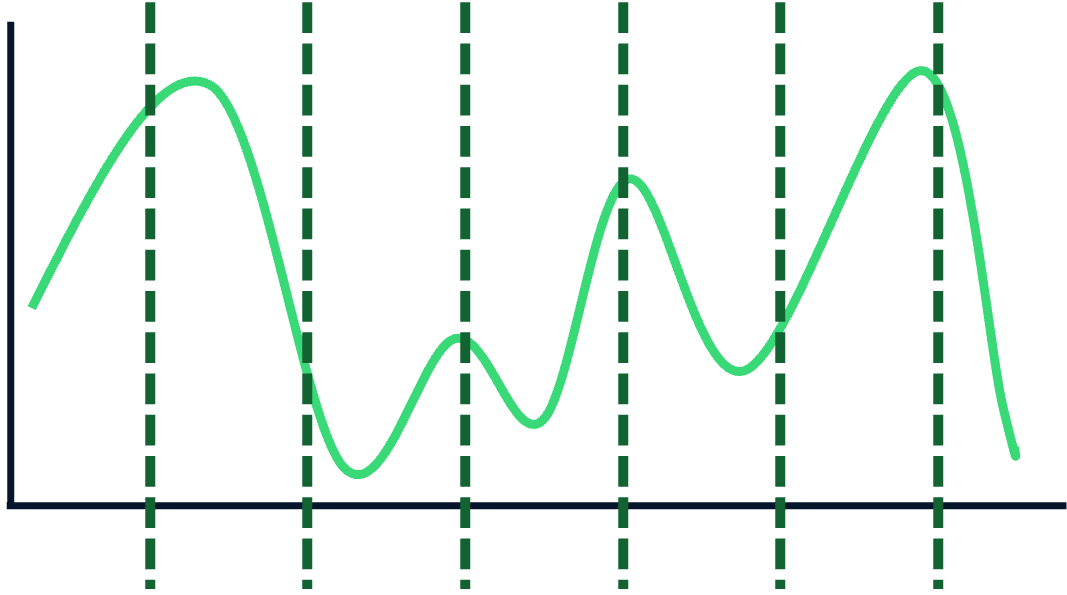}}
        \hfill
     \subfigure[Sliding Window]{
         \centering
         \label{fig:motif-sw}
         \includegraphics[width=0.31\textwidth]{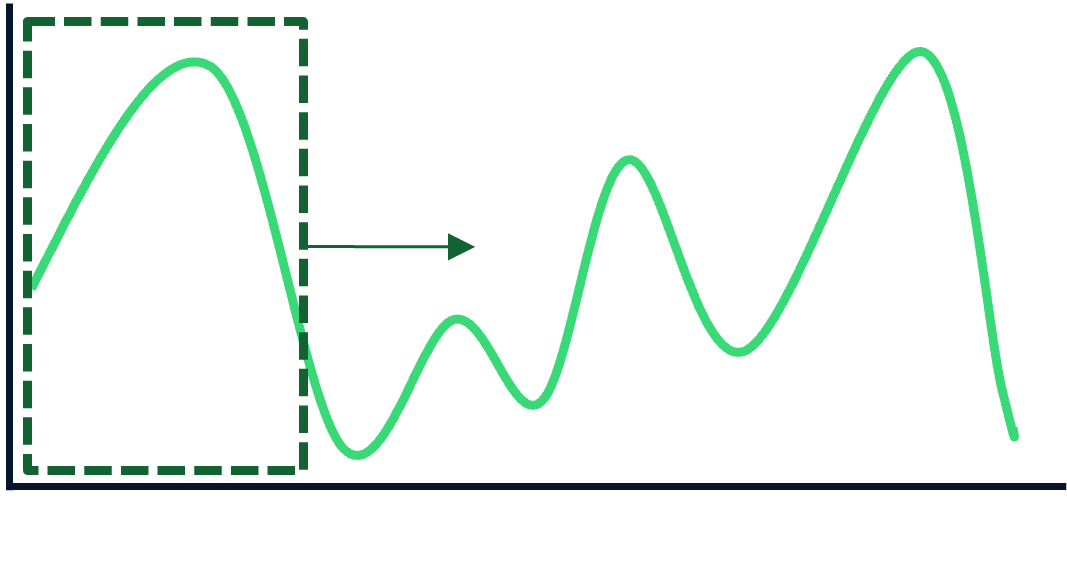}}
     \hfill
     \subfigure[Signal Processing Techniques]{
         \centering
         \label{fig:motif-sp}
         \includegraphics[width=0.31\textwidth]{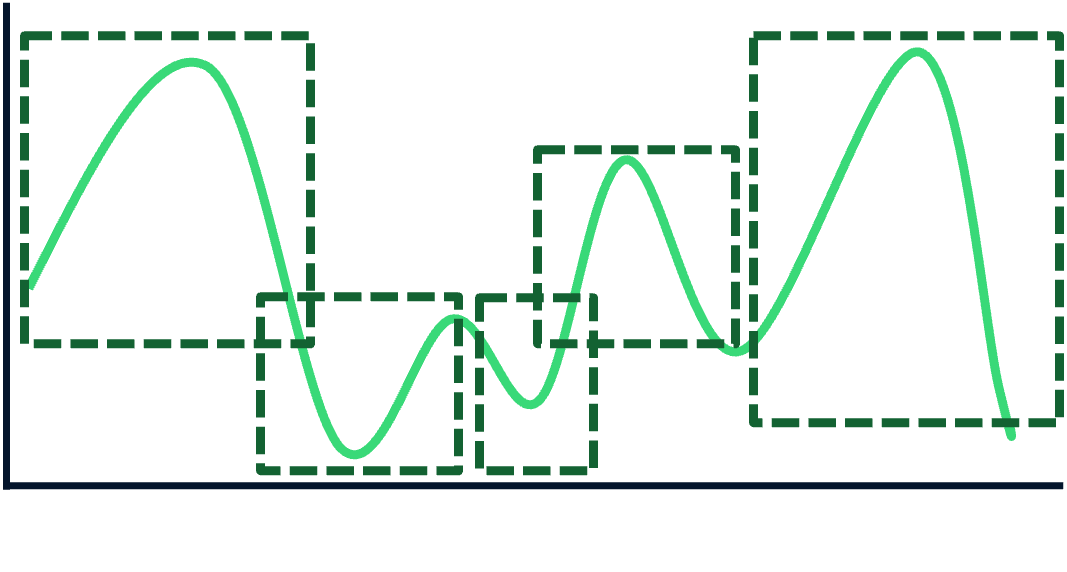}}
    \caption{Example Motif Construction Methods: \subref{fig:motif-chop} chopping the trace into chunks, \subref{fig:motif-sw} using a sliding window, or \subref{fig:motif-sp} using signal processing techniques to automatically identify motifs. 
    % (e.g., using Dynamic Time Warping or Discrete Fourier Transforms).
    }
    \label{fig:motif-construction}
\end{figure*}

\boldsection{Motifs \& Motif Construction}
%%%
% Notation NOTES
% $n$ is the number of time series, $m$ is the number of motifs
% superscript is the *identifier*, subscript is the *time step*
%%%
We first formalize our notion of motifs. A motif is a short segment of the trace which may implicitly capture an underlying behavior within a time series. Real sample traces and motifs are shown in \Autoref{fig:sample-motifs} and Appx.~\ref{sec:appdx-intro}. 
We define a \emph{motif}, $\mu$, %($\mu$, $\sigma$), 
as a short, ordered sequence of values ($v$) 
% from a time series 
of length $\tau$:
\begin{equation}\label{eq:motif-def}
    \mu =[v_i, v_{i+1}, \ldots, v_{i + \tau}]
\end{equation}

We denote a set of $n$ time series traces as $X = [x^1, \ldots, x^n]$. 
% Define a single time series with $t$ timesteps/ has length of $t$ as $x^i = [x_1, x_2, \ldots, x_t]$
Each time series may be represented as a sequence of motifs:
$x^j=[\mu_{1}^{i}, \mu_{2}^{i}, \ldots]$ where each $\mu^{i}_{t}$ gives the motif identifier $i$ at the ordered time step $t$. We also define a motif set $\mathcal{M}$, of $|m|$, which is the
% either from held out dataset or from dataset. 
complete set of motifs generated from the traces $\mathcal{M}=\{\mu^1, \ldots, \mu^m\}$. 
% This depends on the motif generation/construction method.
% but for example in the chopping case, given the motif length $\tau$, the motif set is the union of all size-$\tau$ chunks in the traces. 
% We assume there is a consistent, conclusive way to pull motifs from the traces, and motif discovery is outside the scope of this work. We refer the interested reader to other works focused on this problem \citep{chinpattanakarn2024framework, schafer2022motiflets, ye2009time}. 
We assume there is a consistent, conclusive way to pull motifs from the traces; the user may choose which method they use 
% to pull motifs from the traces 
based on the end application goal.
% The user may choose how they wish to pull out motifs from the time series based on the end application goal. 
Motif discovery is outside the scope of this work since this problem has been widely solved (see \citep{chinpattanakarn2024framework, schafer2022motiflets, ye2009time} and Appx~\ref{sec:appdx-methods}). 
That being said, three straightforward methods to extract motifs from traces, shown in \Autoref{fig:motif-construction}, include chopping the traces into size $\tau$ chunks \subref{fig:motif-chop}, using a sliding window of size $\tau$ to extract motifs \subref{fig:motif-sw}, and using signal processing techniques such as Discrete Fourier Transforms (DFT) to automatically extract motifs \subref{fig:motif-sp}. Our framework is amenable to any motif extraction method.

\boldsection{Granger Causality \& Transfer Entropy}
% First thing to determine - how to define motif causality? Want this notion to be mathemtically concise, and to have a real world meaning / mean something tangible and explainable (e.g., unlike opaque notion used in GlucoSynth of motif caus.)
Granger Causality is a common method to characterize causal relations amongst time series~\citep{granger1969investigating}. Different from other causal methods, Granger defines causal relations in terms of \textit{predictability}. Under Granger Causality, given two time series $x$ and $y$, $x$ causes $y$ if past information about $x$ is more predictive than past information about $y$ only. 
Transfer Entropy (TE), sometimes also called Causation Entropy, is a nonlinear extension of Granger Causality \citep{schreiber2000measuring, barnett2009granger}. Using information theoretic measures, TE measures the amount of uncertainty that is reduced in future states of time series $y$ as a result of knowing the past states of time series $x$. 
% In other words, looks at the reduction in uncertainty about future states of x given the current state of y in addition to x (measure of how much extra information y gives for x).
% TE = Uncertainty reduction of the future states of Y as a result of knowing the past states of X in the context of the world of time series Z
% measures the amount of uncertainty that is reduced in future states of $y$ by knowing past states of $x$ in the context of all the other time series $Z$
Given two time series, $x^i$ and $y^i$, the TE from $x^i$ to $y^i$ is:
\begin{equation}
    TE_{x^i\rightarrow y^i} = H(y^i_t | y^i_{t-1, \ldots, t-f}) - H(y^i_t | y^i_{t-1, \ldots, t-f}, x^i_{t-1, \ldots, t-g}) 
\end{equation}
where $H(\cdot | \cdot)$ is a conditional entropy function and $f$ and $g$ are lag constants.

Traditional TE is under the important assumption that the effect is influenced by the cause under a fixed, constant time delay. However, this assumption does not hold for many real world time series applications, particularly in health streams, where data may be affected by past events at \textit{varying} lengths of time. As such, an extension of TE that allows for different time delays has been developed, denoted here as Variable-lag Transfer Entropy (VTE) \citep{amornbunchornvej2021variable}.
The VTE from $x^i$ to $y^i$ is defined as:
\begin{equation}\label{eq:TE}
    VTE_{x^i\rightarrow y^i} = H(y^i_t | y^i_{t-1, \ldots, t-f}) - H(y^i_t | y^i_{t-1, \ldots, t-f}, x^i_{t-1-\Delta_{t-1}, \ldots, t-g-\Delta_{t-g}}) 
\end{equation}
where 
% $H(\cdot | \cdot)$ is a conditional entropy function, $f$ and $g$ are lag constants and 
$\Delta_t$ is a variable length lag amount.
% $$TE = H(y^i_t | y^i_{t-1, \ldots, t-f}) - H(y^i_t | y^i_{t-1, \ldots, t-f}, x^i_{t-1-\Delta_{t-1}, \ldots, t-\Delta_g}) $$
% where $H(\cdot | \cdot)$ is a conditional entropy function, $f$ is a lag constant and $\Delta_g$ is a variable length lag constant.
% Henceforth, when referring to TE, we are referring to the variable-lag definition of TE.
We will adapt this equation and other notions of TE next for our definition of Motif Causality.
% the succeeding sections

\boldsection{Defining Motif Causality}
Our definition of motif causality is inspired from various Transfer Entropy and Causation Entropy threads (\Autoref{eq:TE}, \citet{irribarra2024multi, gong2023causal,  amornbunchornvej2021variable, assaad2021mixed, sun2015causal}).
% Generalize / Specify TE to a set of motifs and allow for variable length lags
% Combining threads related to multivariate TE, variable lag TE and with/for motifs.
% causation entropy paper - with multivariate TE ~\cite{sun2015causal}
% Variable lag te paper ~\cite{amornbunchornvej2021variable}
% temporal causation entropy which has selection of most optimal lag constants ~\cite{assaad2021mixed}
The Motif Causality (MC) from motif $\mu^i$ to motif $\mu^j$ conditioned on the set of motifs $\mathcal{K}$ is defined as:
\begin{equation}\label{eq:motif-caus}
    MC_{\mu^i\rightarrow\mu^j|\mathcal{K}} = 
    H(\mu^j_t | \mathcal{K}_{t-1, \ldots, t-f}) - 
    H(\mu^j_t | \mathcal{K}_{t-1, \ldots, t-f}, \mu^i_{t-1-\Delta_{t-1}, \ldots, t-g-\Delta_{t-g}})
\end{equation}
where $\mathcal{K}\subset\mathcal{M}$, 
% the global motif set $\mathcal{M}$, 
$H(\cdot | \cdot)$ is a conditional entropy function, $f$ and $g$ are lag constants and 
$\Delta_t$ is a variable length lag amount. 
Essentially, this provides a measure of information gain by determining how much uncertainty is reduced for $\mu^j$ by 
% observing the past states of $\mu^j$ 
observing past occurrences of $\mu^i$
compared to the ``status quo", the set of the rest of the motifs $\mathcal{K}$. 
The $MC$ value will be between 0.0 and 1.0. Higher values indicate stronger causality (i.e., more uncertainty about the future is reduced for $\mu^j$ given $\mu^i$).

\boldsection{Conditional Entropy Function}
To implement the conditional entropy function, $H(\cdot | \cdot)$, there are many different types of entropy which the user can choose based on their end goal or application. For our purposes, we elucidate two common ones: Shannon entropy \citep{shannon1948mathematical} and R\'enyi entropy \citep{jizba2012renyi, jizba2022causal}.
Shannon entropy is defined as:
\begin{equation}
    H(x^i) = - \sum_t{p(x^i_t)\log_2(p(x^i_t))}
\end{equation}
where $p$ is a probability distribution.
R\'enyi is more flexible at estimating uncertainty and defined as:
\begin{equation}
    H_\alpha(x^i) = \frac{1}{1-\alpha}\log\left(\sum^n_{v=1}{p_v^\alpha}(x^i)\right)
\end{equation}
where $\alpha$ is a weight parameter, $\alpha>0$. When $\alpha\rightarrow 1$ R\'enyi entropy converges to Shannon entropy. We note that there are many other types of entropy functions that could be used and might be relevant, such as Wavelet \citep{rosso2001wavelet} or Permutation Entropy \citep{bandt2002permutation}.

\begin{figure}[t]
    \centering
    \includegraphics[width=0.9\linewidth]{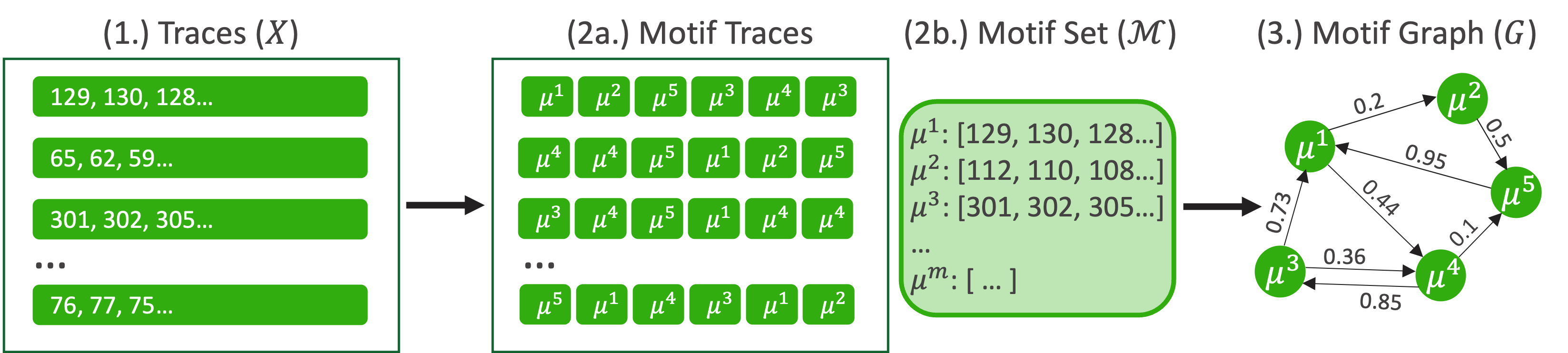}
    \caption{Preprocessing Steps. $X$ are transformed to motif traces and $\mathcal{M}$ and then into a graph $G$.}
    \label{fig:preproc}
\end{figure}

%% Part 2: Learn Motif Causality
\section{Motif Causal Discovery Framework}\label{sec:learn-MC}
% Now that we have a formal definition of motif causality, we next describe how we learn motif causality starting from a set of time series traces, including generating the motifs and motif set, and learning the causal relationships amongst motifs.
Now that we have formulated our definition of Motif Causality, we next describe our casual discovery framework \MD to learn motif causal relationships amongst a set of time series motifs. We first detail  preliminaries related to the problem definition and preprocessing in \Autoref{sec:appr-prelim}, describe the model architecture in \Autoref{sec:model-arch} and finish with the model training algorithm in \Autoref{sec:model-train}.

% \begin{wrapfigure}{r}{0.3\textwidth}
%     \centering
%     \includegraphics[width=0.3\textwidth]{Figures/motif_causal_graph.png}
%     \caption[width=0.3\textwidth]{Example Motif Causal Graph}
%     \label{fig:motif-graph}
% \end{wrapfigure}

\subsection{Preliminaries}\label{sec:appr-prelim}

\boldsection{Problem Definition}
Since we do not know nor have any way to determine the underlying motif-causal structure (i.e., we have no ground truth), we formulate this problem as an unsupervised graph link prediction problem:
Given a set of input time series motifs extracted from our set of traces $X$, and the complete set of nodes which is equivalent to the motif set $\mathcal{M}$ (i.e., each motif is a node), predict the edges between all the nodes. In other words, learn the edge weights, the motif causal relationships, between all the nodes, the motifs. 
% output a motif causal graph $G$, that represents all the motif causal relationships between all of the motifs, extracted from the input time series. 
% A simple example graph is shown in \Autoref{fig:motif-graph}.
To solve this problem, we build a Graph Neural Network-based causal discovery framework that learns the MC edge weights in an unsupervised manner. We walk through each part of the framework next, starting with the preprocessing steps.

\boldsection{Preprocessing}
An overview is shown in \Autoref{fig:preproc}. Motifs of length $\tau$ are pulled from the input time series  to create a set or ordered motif traces and the motif set $\mathcal{M}$ 
% with a total of $m$ motifs 
(see \Autoref{sec:define-motif-caus}). 
% For purposes of this running example 
In our implementation, we use the chopping method to generate motifs; 
% Additionally, the motif set $\mathcal{M}$ with a total of $m$ motifs is also generated from the traces. 
% Since we used the chopping method, given the motif length $\tau$, the motif set 
$\mathcal{M}$ is the union of all size-$\tau$ chunks in the traces. 
% Motif dictionary is a dictionary that tracks each motif identifier and the actual motif values (e.g., \{1: \[100, 121, 122, 123\]\}. 
% For practicality, we implement the motif set as a dictionary. The dictionary contains the motif identifiers and the actual time series values contained in each motif (which are important for later algorithm steps.) The keys of the dictionary are generated by hashing the motif values for quick lookup and retrieval of the motif identifiers and their location in the dictionary. 
From there, an initial motif graph is generated from the motif traces. 
% An example motif causal graph is shown in \Autoref{fig:preproc} (3).
In the graph structure, each node contains the motif identifier and the actual motif values (e.g., motif $\mu^1 = [129, 130, 128, ...]$). Edges represent directed motif-causal relationships between two motifs. The edge weight indicates the \textit{strength} of the causal relationship.
At this stage,
generating the \textit{best} graph is not our primary concern since the framework will add and remove optimal edges during the learning process,
% (that is what the learning will be for), 
and we just need a starting graph structure to build from. 
As such, we believe it an acceptable first pass to assume 
% As such, we believe it is acceptable for a first pass to assume 
there is \textit{some} degree of causality between motifs that appear immediately one after the other, and build the preliminary graph 
% according to this assumption. 
% We generate the initial graph 
by adding edges between each subsequent motif in the traces. For example, for the first motif trace in Fig.~\ref{fig:preproc} (2a), edges would be added from $\mu^1\rightarrow\mu^2$, $\mu^2\rightarrow\mu^5$, 
% $\mu^5\rightarrow\mu^3$, 
etc. 
We compute the MC edge weight 
% as the MC between the motif time series values 
according to \Autoref{eq:motif-caus}.

\begin{figure}[t]
    \centering
    \includegraphics[width=\linewidth]{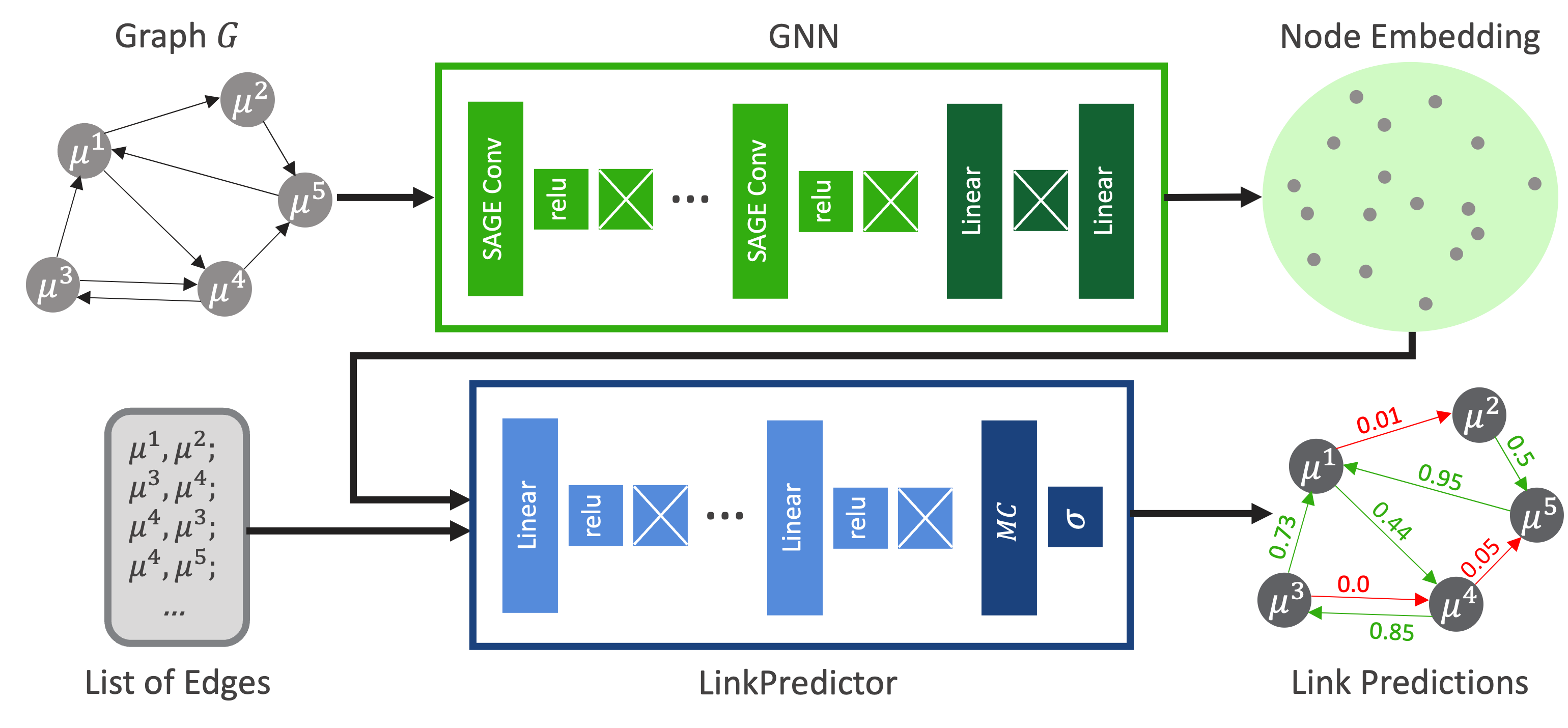}
    \caption{Model Architecture: a GNN consisting of stacks of GraphSAGE, ReLU, Dropout (``X" boxes) and Linear layers learns a node embedding. The LinkPredictor consists of Linear, ReLU and Dropout layers followed by the ``MC" motif causality layer and a Sigmoid layer ($\sigma$ box). The LinkPredictor takes in the node embedding and outputs the predicted edges and their weights.}
    \label{fig:model-arch}
\end{figure}

\subsection{Model Architecture}\label{sec:model-arch}
An overview of the model architecture is shown in \Autoref{fig:model-arch}. The model consists of two main components: a Graph Neural Network (GNN) 
% with GraphSAGE layers 
that uses GraphSAGE layers to learn node embeddings, and a LinkPredictor network that uses the learned node embeddings to predict motif causal links amongst nodes. We detail each  component next and then describe model training in \Autoref{sec:model-arch}.

\boldsection{GNN} The Graph Neural Network takes in a starting motif graph $G$, and outputs a learned node embedding. The GNN is structured using GraphSage convolutional layers. 
% The GNN uses GraphSAGE as its underlying layers.
GraphSAGE (standing for Sample and Aggregate)~\citep{hamilton2017inductive} learns low dimensional vector representations of nodes. The core intuition of GraphSAGE is that a node is known by the company it keeps (i.e., its neighbors). The algorithm works by iterating over a sample of the node's neighboring nodes and ``aggregating" their embeddings in order to determine the current node's embedding. 
% As in many Graph Representation Learning algorithms, the idea is for the GNN to optimize the mapping such that nearby connected nodes in the original graph structure also remain close within the embedding space, and unconnected nodes are shoved far apart in the embedded space. In this way, graphical relationships from the original network remain preserved inside the embedded space.
Importantly, GraphSAGE is \textit{inductive}, meaning it can generalize to unseen nodes.  GraphSage use both node features and topological structure (i.e., the graph structure) of each node’s neighbourhood simultaneously to efficiently generate representations for new nodes without requiring model retraining. To do this, the algorithm relies on its aggregation function, also known as the message-passing process, which learns how to aggregate node features based on encoding information about a node's local neighborhood. Therefore, when given new node data, the function uses the local neighborhood of the node to aggregate the features appropriately, and learn the embedded feature representation (as opposed to needing to learn a unique embedding for every individual node).
% rest of the architecture
As shown in \Autoref{fig:model-arch}, in our framework the GNN consists of sequential stacks of GraphSAGE convolutional, ReLU and dropout layers (represented by the ``X" boxes in the figure) followed by the message-passing layers-- stacks of linear and dropout layers. 
The motif time series values are the node data used to learn the embeddings.
% The node data used to learn the embeddings are the motif time series values themselves. 
We instantiate the aggregation function using mean aggregation for simplicity.

\boldsection{Link Predictor}
The LinkPredictor network takes in learned node embeddings outputted from the GNN and a list of edges to predict, and returns the probability of each edge and the motif causality values for each edge. 
% The LinkPredictor computes the Motif Causal values between the two nodes, following \Autoref{eq:motif-caus}. 
At its core, the LinkPredictor learns a function to predict the probability of an edge between two nodes. 
To do this, it computes the MC between the nodes using \Autoref{eq:motif-caus}, 
and a probability score, represented by the element-wise dot product of the two embedded node vectors. 
To implement the MC computation, we adapt existing 
Transfer Entropy libraries and compute the conditional entropy function, $H(\mu)$, using the histogram method \citep{behrendt2019rtransferentropy}; additional details in Appx.~\ref{sec:appdx-methods}.
% The LinkPredictor learns a function to compute the probability of an edge between two nodes given the nodes' two embedding vectors and the computed Motif Causality value between the two nodes. 
% This is important because it 
In terms of architecture, the LinkPredictor is implemented via stacks of linear, ReLU and dropout layers followed by Motif Causality ($MC$) and sigmoid layers to learn the edge probability function.
% , tempered by the motif causal values. 
% To compute the probability about whether or not there exists a link between two nodes, 
The network balances evaluating the product of the embedded node vectors and the motif causal values between the motifs themselves,
% the network learns a function which balances optimizing the product of the embedded vectors, and maximizing the motif causal values of the motifs (the node features) themselves.
allowing the network to learn how to optimize edge addition/deletion in the graph guided by the underlying causality. (More details provided next in \Autoref{sec:model-train}).
In this way, the 
LinkPredictor learns to predict edges that have high motif causality
with high probability.
% why do both ? 
% The LinkPredictor computes \textit{both} the probability function and the Motif Causality values because ...
% Link pred learns to correctly predict where edges are in the graph, guided by the MC calculations. 
% The probability score is represented by the element-wise dot product of the two embedded vectors. 
% MC is computed following \Autoref{eq:motif-caus}. 
% 
% A simplified depiction is shown in \Autoref{fig:TE-hist} in Appx.\ref{sec:appdx-methods}. 

% To implement the MC computation, we adapt existing Transfer Entropy libraries \citep{behrendt2019rtransferentropy} and compute the conditional entropy function (i.e., $H(\mu)$, Shannon's or R\'enyi entropy) using the histogram method. A simplified depiction is shown in \Autoref{fig:TE-hist} in \ref{sec:appdx-methods}. Essentially, motif time series values are binned into a histogram.
% % bins.
% % are binned based on their values into histogram bins. 
% % From there, 
% Distributions between motifs can be compared to determine how much uncertainty about future predictions of the motif is reduced in the distribution. For example, in \ref{fig:big-red} when computing MC for $\mu^i$, $\mu^j$ covers a larger distribution, resulting in a large reduction in uncertainty and higher motif causality, whereas \ref{fig:no-red} $\mu^z$ covers hardly any new distribution compared to the status quo ($H(\mu^i|\mathcal{K})$), resulting in a small reduction in uncertainty and low causality for $\mu^z$.

\begin{algorithm}[t]
\caption{Training Procedure to Learn Motif Causality}
\label{alg-learn-motif-caus}
\SetAlgoLined
\DontPrintSemicolon
\SetNoFillComment
\SetKwFunction{GNN}{\textbf{GNN}}
\SetKwFunction{LP}{\textbf{LinkPredictor}}
\SetKwFunction{negSamp}{\textit{negative\_sample\_edges}}
\SetKwFunction{samp}{\textit{sample}}
\SetKwFunction{update}{\textit{update\_edges}}
\KwIn{Input Graph $G$, Epochs $e$, Edge Prediction Threshold $\theta$}
% \KwOut{output is}
% edge\_index $\longleftarrow$ $G$.edges \;
% motifs $\longleftarrow$ $G$.data \;
\For{e epochs}{
    \tcc{Compute Node Embedding}
    node\_emb = \GNN($G$)\;
    \tcc{Get predictions on positive edges}
    %this is list of positive edges - 2d array of nodes 
    $E$ $\longleftarrow$ \samp($G.edges$)\tcp{Edges that exist in $G$}
    $p$, $c$ = \LP(node\_emb, $E$)\;
    \tcc{Get predictions on negative edges}
    $\hat{E}$ $\longleftarrow$ \negSamp{G}\tcp{Edges not in $G$}
    $\hat{p}$, $\hat{c}$ = \LP(node\_emb, $\hat{E}$)\;
    \tcc{Compute Loss}
    $y = \gamma \times p + \lambda \times c$\;
    $\hat{y} = 1 - (\gamma \times \hat{p} + \lambda \times \hat{c})$\;
    $loss = - \sum_{i=1}^b \left(y_i \log \hat{y}_{i} + (1-y_i)\log(1 - \hat{y}_{i}) \right)$\;
    \tcc{Update edges in the graph based on new predictions}
    $G$.\update($E, y, \hat{E}, \hat{y}$, $\theta$)\;
}
% \Return{node\_emb, linkPredictor}
\end{algorithm}
% Minimizing Log Likelihood Loss:
% \begin{equation}
%     l(\theta) = - \sum_{i=1}^n \left(y_i \log \hat{y}_{\theta,i} + (1-y_i)\log(1 - \hat{y}_{\theta,i}) \right)
% \end{equation}
% $\hat{y}$ is the predicted probability of the positive class, $y$ is true labels
% Log loss works by only counting prediction values/probabilities associated with the true labels

\subsection{Model Training}\label{sec:model-train}
% Now that we have an understanding of the model architecture, 
We next describe how the entire framework is trained together, shown in \Autoref{alg-learn-motif-caus}. 
First, the graph is fed through the GNN network to learn an embedded representation of the nodes (Line 2). 
Next, the LinkPredictor makes predictions on a sample of the edges that exist in the graph $G$, denoted as the positive edges $E$ using the learned node embedding (Line 3-4).
Then, a batch of edges that do not exist in $G$, 
% (i.e., nodes with no connections between them), denoted as \textit{negative edges}, 
$\hat{E}$, are randomly sampled by selecting random pairs of nodes with no connections between them 
% that do not exist in $G.edges$ 
(Line 5). The LinkPredictor makes predictions on the negative edges (Line 6). 

From there, the positive predictions $y$ are computed by combining the positive edge predictions $p$ and the motif causality values for the positive links $c$ (Line 7).
Similarly, the negative predictions $\hat{y}$ are computed by combining the negative link predictions $\hat{p}$ with the MC values for the negative links $\hat{c}$ (Line 8). $\gamma$ and $\lambda$ are important hyper-parameters which balance the influence of the dot product predictions vs. the motif causality. The model is trained to minimize the log likelihood loss function, computed following the equation on Line 9. 
Essentially, this loss function optimizes the model to maximize its predictions of positive edges (true links, edges that have high motif causality values) and minimize predictions of negative edges (links that should not exist in $G$, edges with low MC).
Finally, the edges in $G$ are updated based on the model edge predictions 
% predicted positive and negative edges 
and an edge threshold $\theta$ (Line 10). $\theta$ is a user-specified parameter with a range between 0 and 1. If any of the edges in $\hat{y} \geq \theta$, they are added to the graph $G$; if any of the edges in $y < \theta$, they are removed from $G$.

Due to the dual training between the GNN and the LinkPredictor, as the model iterates the node embeddings are continuously updated based on new linkages that may be added or removed. As such, the GNN learns good node embeddings representative of the types of nodes that are close to each other and have connections to each other (nodes that are very motif causal of one another). 
% GNN gets better and better at embedding nodes motif causal of each other near each other (with links). 
% The Link Predictor improves its prediction function about where nodes occur based on the embedding, guided by motif causal values. 
The LinkPredictor is guided by the motif causality values between motifs to help it predict where linkages should be, and as it 
learns characteristics of highly causal edges, and gets better and better node embeddings from the GNN, it learns a good prediction function for predicting node linkages.

\boldsection{Making New Edge Predictions} To make a link prediction between two new nodes using the trained framework, the nodes are first fed through the GNN to get their node embeddings, and then the embeddings are sent through the trained LinkPredictor network, which returns the predicted probability they have an edge between them, along with the motif causality value.

%% Part 3: Use Motif Causality for other stuff
\section{Uses of Learned Motif Causality}\label{sec:use-cases}
% To illustrate the use of Motif Causality as a building block for downstream tasks, here we integrate MC with three model use cases of Forecasting, Anomaly Detection and Clustering, and evaluate their performance later in \Autoref{sec:eval}. 
To illustrate the use of Motif Causality as a building block for other downstream tasks, in this section we integrate Motif Causality with three example model use cases of Forecasting, Anomaly Detection and Clustering. 
% Graphical depictions of each of these implementations are available in \ref{sec:appdx-methods}.
We evaluate the performance of these use cases later in \Autoref{sec:eval}. 
% We note, however, that motif causality may be usable for a variety of other tasks as well.
% All of these models integrate with the trained motif causal graph outputted from training the Motif Causality Framework (and detailed previously in \Autoref{sec:learn-MC}).

\begin{figure}[t]
    \centering
    \includegraphics[width=0.65\linewidth]{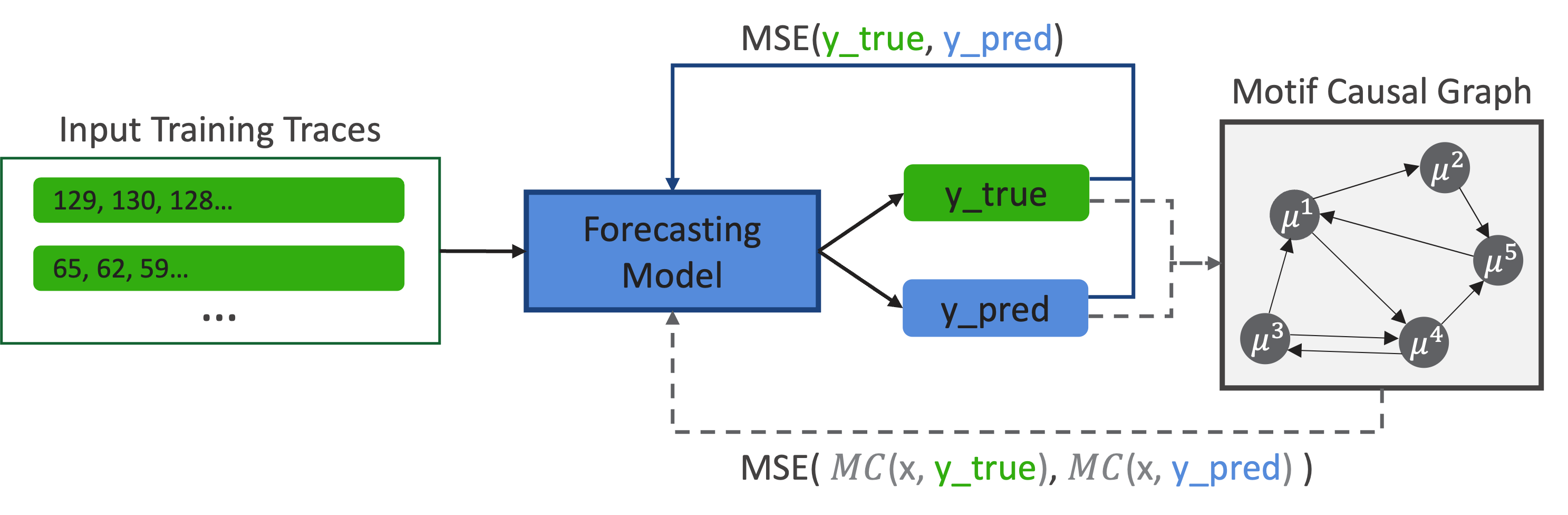}
    \caption{Forecasting Model integrated with Motif Causality.}
    \label{fig:forecasting}
\end{figure}

\begin{figure}[t]
    \centering
    \includegraphics[width=0.65\linewidth]{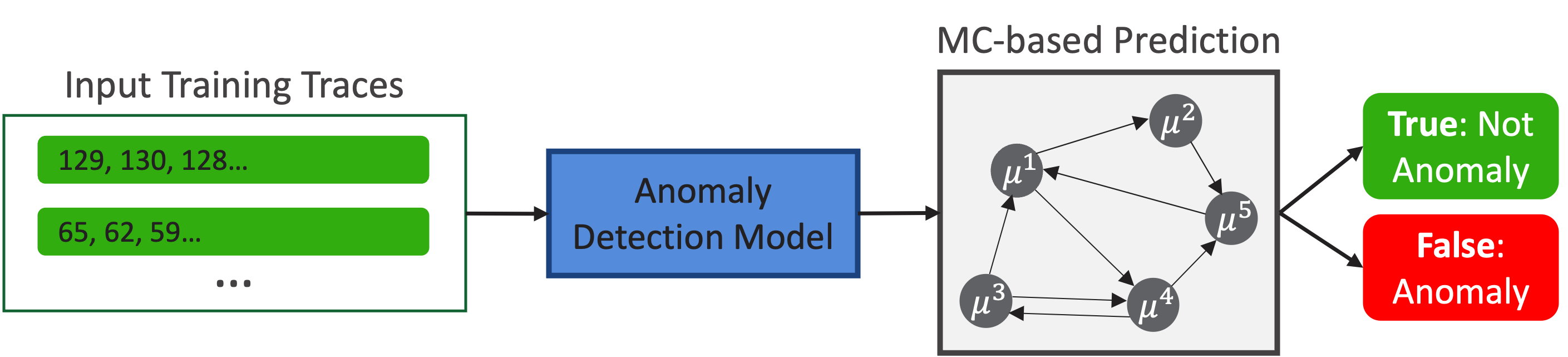}
    \caption{Anomaly Detection Model integrated with Motif Causality.}
    \label{fig:anomaly}
\end{figure}

\boldsection{Forecasting}
Our integration is shown in \Autoref{fig:forecasting}. 
Briefly, some basic forecasting models work by sliding a window across input traces to learn sequential time series patterns. 
% taking in a sliding window along the input traces to learn sequential time series patterns. 
Their loss function computes the difference between the model's predicted future timesteps (y\_pred) and the ground truth future time steps (y\_true) in the traces.
To integrate motif causality, we add an additional loss function that seeks to minimize the difference in the MC between the previous time step ($x$) and 
the predicted future time steps (y\_pred) vs. 
$x$ and
the ground truth future time steps (y\_true). 
The intuition is that there should be similar causality between the previous time step and the predicted future time step as the ground truth data.
% predicted future time series should have similar motif causal values from the previous  as the ground truth data.
MC is computed using the trained motif causal graph outputted from the MC framework.
We assume the motif size $\tau$ is the same as the forecasted prediction window size to ensure MC is computed on comparable time chunks (i.e., motifs).
% This is important so that causality between motifs can be pulled from the predictions (predicted motifs) correctly.

\boldsection{Anomaly Detection}
Integration of a basic anomaly detection model with MC is shown in \Autoref{fig:anomaly}. 
We add an MC-based anomaly prediction block after the model that uses the predicted MC between  previous timesteps and future timesteps to determine if the next trace chunk may be anomalous or not.
Specifically, it checks if the MC between the previous time steps (previous motif) and the next one are less than a threshold, and if so classifies it as an anomaly.
The size of the predicted time chunk must be the same size as the motif size $\tau$ and we suggest the anomaly threshold be set to the edge prediction threshold $\theta$, since this was what was used to train the motif causal graph originally.

\boldsection{Clustering}
For clustering, there is typically a method to group clusters based on minimizing distances between each data point and the cluster centroid. These distances are computed using various distance measures such as 
% euclidean or manhattan distance, or for time series using time series metrics like 
Discrete Time Warping (DTW) \citep{sakoe1978dynamic} for traces. To integrate MC with a basic clustering algorithm, 
% when computing the distance metric we also compute the Motif causality values 
we use the motif causality values as an additional distance metric. 
% to compute distances between data points and centroids.
% as well as other data points 
% are computed in addition to the normal distance metric. 
The intuition here is to add an element of \textit{causality} to the clustering, such that as the algorithm learns, MC values within each cluster will be minimized and similar data points within the cluster should be \textit{motif causal} of each other. An example is shown in \Autoref{fig:clustering} in Appx.~\ref{sec:appdx-methods}.
% : the MC value between the blue centroid and the blue data point to the right is high with 0.9, whereas the MC between the blue centroid and the green data point belonging to a different cluster is low at 0.11.

% Assign clusters - normally computes distance between each data point and the cluster centroid, tries to minmize distance and assigns each data point to the closest cluster

% Computes distance using a time series based metric, specifically DTW distance

% In MC version, computes DTW distance + MC value.
% Trying to also optimize causality values in the clusters, so the clustering could have some real world meaning related to causality --> motifs in similar clusters also share similar causality characteristics

%% file: 4.eval.tex
\section{Evaluation}\label{sec:eval}

\begin{wrapfigure}{r}{0.4\textwidth}
\vspace{-1.9cm}
    \centering
    \includegraphics[width=0.4\textwidth]{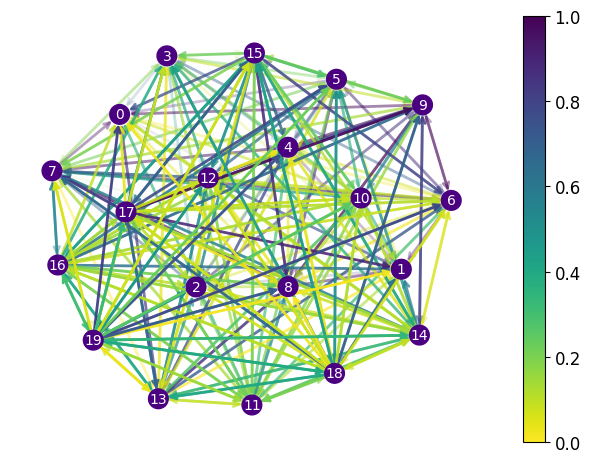}
    \caption{Learned Motif Causal Graph for $n=10$, $|\mathcal{M}| = 20$. Edge color indicates the MC value.}
    \label{fig:MC-graph}
\vspace{-1.0cm}
\end{wrapfigure}
As mentioned in \Autoref{sec:rel-work}, 
% there is no prior work on quantifying causality amongst motifs, and 
existing Granger causal techniques are not directly comparable with \MD due to their framework set-up or model assumptions; i.e., they formulate causality using multivariate statistical properties or temporal statistical measures using time lags which do not hold for motifs, or require labels or a known underlying causal structure which are not available for our health data streams (see Appx.~\ref{sec:appdx-eval}.)
% Additional details are available in Appx.~\ref{sec:appdx-eval}. 
Therefore, in this section we evaluate our causal discovery framework in terms of scalability and performance for three downstream use cases: forecasting, anomaly detection and clustering. 

% Specifically, they have one or more of the following issues: They compose relations between two time series, and cannot find relationships amongst a set of motifs or traces \cite{gong2023causal, amornbunchornvej2021variable}; they formulate causality based on various statistical properties amongst variables- things like variable-based correlation and density-, classification- and prediction-based error measures between variables
% \cite{amornbunchornvej2021variable, bonetti2024causal, irribarra2024multi}. These assumptions do not hold for univariate time series motifs.
% Compose causal relationships 
% over time series using repeated statistical measures or temporal dynamics – things like learning implicit repeated patterns, or computing sufficiency or faithfulness hypotheses over lags across the full time series \cite{sun2015causal, assaad2021mixed, lowe2022amortized, najafi2023entropy, lamp2024glucosynth, pan2024effcause, tank2021neural}
% These methods do not work for motifs because they are short time series (so there would be no repeated patterns, nor will statistical hypotheses like sufficiency or faithfulness hold since they are not evaluating multiple time lags repeated over a trace.).

% Require labels or a known underlying causal structure to guide model training \cite{gong2023causal, najafi2023entropy, bonetti2024causal}, which is not available for our data and many similar event-based data streams.

\paragraph{Experimental Details.}
We use three health stream datasets with different lengths and sampling frequencies: glucose traces from continuous glucose monitors~\cite{akturk2021real}, ECG traces and respiration (Resp) traces from the MIMIC-BP dataset~\cite{dvn_2023}. All experiments were run 5 times
with an 80/20\% train/test split
on an Intel Sky Lake 48 CPU VM with 192GB of RAM. Motifs were pulled from traces using the chopping method and we set $\gamma=0.7$ and $\lambda=0.5$. Additional details are available in Appx.~\ref{sec:appdx-eval}.

%%%%%%%%%%%%
% Add note here that no ground truth or other versions for comparison so instead do scalability + external validation with use cases 

\boldsection{Motif Causal Graphs}
An example motif causal graph from the \MD framework is shown in \Autoref{fig:MC-graph}. The edge color indicates the strength of the MC relationship; darker is stronger (closer to 1) while lighter is less strong (closer to 0). 
% \Autoref{fig:MC-examples-strong},~\ref{fig:MC-examples-low} in 
\ref{sec:apdx-mcfigs} shows example MC between motifs of different $\tau$.
% Example graphs for 10 traces, trained for 10 epochs
% Darker values indicate stronger causality relationships

% \begin{figure}[t]
%     \centering
%     \includegraphics[width=0.5\linewidth]{Figures/LinkPredGraph_nseq10_tau120_epochs10_theta0.png}
%     \vspace{-0.5cm}
%     \caption{Learned Motif Causal Graph for $n=10$, $|\mathcal{M}| = 20$. Edge color indicates the MC value.}
%     \label{fig:MC-graph}
% \end{figure}

% \begin{figure}
% % \begin{wrapfigure}{r}{0.32\textwidth}
%      \centering
%       \subfigure[$n$ vs. Training Time]{
%          \centering
%          \label{fig:scale-trace}
%          \includegraphics[width=0.32\textwidth]{Figures/VaryTraceLengths.png}}
%         \hfill
%      \subfigure[$\tau$ vs. Training Time]{
%          \centering
%          \label{fig:scale-motif}
%          \includegraphics[width=0.32\textwidth]{Figures/VaryMotifLengths.png}}
%          \hfill
%      \subfigure[$|\mathcal{M}|$ vs. Training Time]{
%          \centering
%          \label{fig:scale-motifset}
%          \includegraphics[width=0.32\textwidth]{Figures/VaryMotifSetSizes.png}}
%     \caption{Scalability varying \subref{fig:scale-trace} \# of traces ($n$), \subref{fig:scale-motif} motif lengths ($\tau$)  and \subref{fig:scale-motifset} motif set sizes ($|\mathcal{M}|$).}
%     \label{fig:scalability}
% \end{figure}

\begin{figure}[t]
\vspace{-0.5cm}
% \begin{wrapfigure}{r}{0.32\textwidth}
     \centering
      \subfigure[$n$ vs. Training Time]{
         \centering
         \label{fig:scale-trace}
         \includegraphics[width=0.32\textwidth]{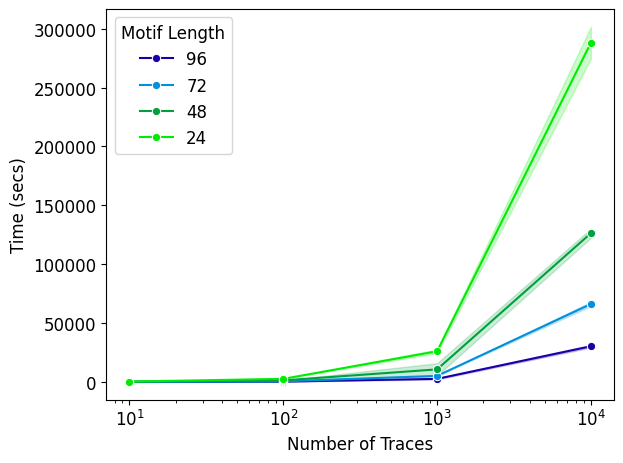}}
        \hfill
     \subfigure[$\tau$ vs. Training Time]{
         \centering
         \label{fig:scale-motif}
         \includegraphics[width=0.32\textwidth]{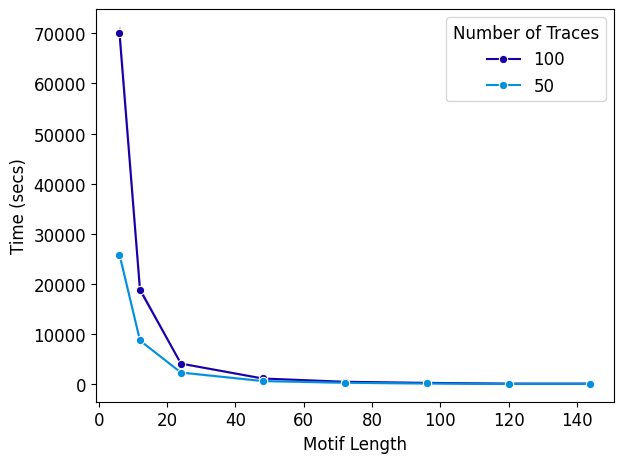}}
         \hfill
     \subfigure[$|\mathcal{M}|$ vs. Training Time]{
         \centering
         \label{fig:scale-motifset}
         \includegraphics[width=0.32\textwidth]{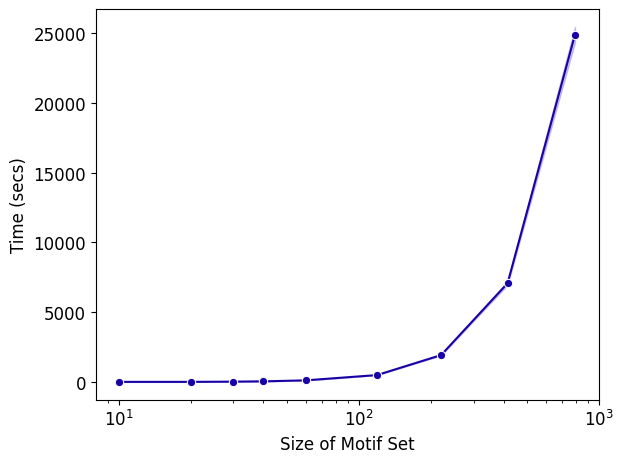}}
    \caption{Scalability for glucose traces varying \subref{fig:scale-trace} \# of traces ($n$), \subref{fig:scale-motif} motif lengths ($\tau$)  and \subref{fig:scale-motifset} motif set sizes ($|\mathcal{M}|$). Results for the other data streams are available in Appx.~\ref{sec:appdx-scale}.}
    \label{fig:scalability}
\end{figure}

\boldsection{Scalability}
Scalability evaluation is shown in \Autoref{fig:scalability} and Appx.~\ref{sec:appdx-scale}, computing total time to train the \MD causal discovery framework in seconds for (a) different numbers of traces $n$, (b) motif lengths $\tau$ and (c) motif set sizes $|\mathcal{M}|$. 
% \ref{fig:scale-trace} different numbers of traces ($n$), \ref{fig:scale-motif} motif lengths ($\tau$) and \ref{fig:scale-motifset} motif set sizes ($|\mathcal{M}|$). 
% the model was trained for 10 epochs. 
Training was for 10 epochs in all experiments.
% and results were averaged across 5 runs.
% \Autoref{fig:scale-trace} plots the training time for different numbers of traces at motif lengths of 24, 48, 72 and 96. 
% \Autoref{fig:scale-motif} plots training time across different motif lengths ($\tau$) for 100 and 50 traces. \Autoref{fig:scale-motifset} plots training time compared to different sizes of the motif set $|\mathcal{M}|$ for 10 traces. 
As to be expected, training time increases for larger $n$ and $|\mathcal{M}|$.
Interestingly, time reduces as the $\tau$ increases, which may in part be because larger $\tau$s mean there are less total motifs (i.e., $|\mathcal{M}|$ is smaller). 
% The time in particular jumps up for $\tau < 10$.

% Compare with GlucoSynth because they have a notion of motif causality. However, since the end goal is completely different (GS used for synthesizing high fiedelity synthetic time series traces, not causl discovery) no other parts of the eval are directly comparable.

% Might make sense to move details of the actual comparisons here or the implementation details in the methods to the appendix .....
\boldsection{Use Cases}
We next evaluate the suitability of Motif Causality to help in three downstream tasks: Forecasting, Anomaly Detection and Clustering. For each use case, we build a simple base model and integrate MC following the architecture descriptions in \Autoref{sec:use-cases}. 
We then compute a set of evaluation metrics to compare the performance between the base model and the MC integrated one. 
% Complete results are reported in \Autoref{table:use-case-eval}. 
% In each use case, both networks are trained for the same number of epochs and with the same base model architecture to ensure a fair comparison.
% For use cases, given each architecture described in \Autoref{}, train a non motif causal version, a normal version and then train one that uses motif causality. 
% For each use case both networks are trained for the same number of epochs to ensure a fair comparison. Report a set of metrics for each use case to evaluate how well it does compared to the normal one.

\boldsection{Forecasting}
For the forecasting task, we use a simple bidirectional LSTM as our base model. 
% To ensure applicability of the motif causal graph (i.e., trace chunks being compared are in the same dimensional space and of same lengths), 
We set the sliding window size, $\tau$ and forecasting window to 6 timesteps.
% (corresponding to 30 minutes of time).
% We train the causal discovery framework for 20 epochs and the forecasting models for 2000 epochs. 
We train the causal discovery and forecasting models for 20 and 2000 epochs, respectively. 
The Root Mean Square Error (RMSE) is reported in \Autoref{table:use-case-eval}; the MC model outperforms the base one for all datasets.

\boldsection{Anomaly Detection}
% Since in many use cases there are no training labels, we use an unsupervised anomaly detection method and build an autoencoder.
For this task, we build a simple autoencoder consisting of stacks of sequential dense layers. 
In the base model, an anomaly is detected if the Mean Absolute Error (MAE) of the reconstructed data (i.e., the trace fed through the encoder and then returned via the decoder) is less than a reconstruction threshold
% . We set the reconstruction threshold to be 
which we set as the standard deviation of the mean of the normal (non-anomalous) training data.
In the MC-integrated version, we detect an anomaly if the predicted MC value between the previous time chunk (motif) and the current one is less than the edge prediction threshold $\theta$. 
We set the sliding window size and $\tau$ to 48, $\theta$ to 0.1 and train the causal discovery and anomaly models for 10 and 50 epochs, respectively.
% $\theta$ is kept the same for training the causal discovery framework and anomaly detection, and set to 0.10 for this experiment.
% The sliding window size used to detect anomalies is the same as the motif size ($\tau$) and set to 48. Causal discovery model was trained for 10 epochs and both anomaly models were trained for 50 epochs.
We compare the models by computing a set of classification metrics including accuracy, F1, Sensitivity and Specificity using the ground truth labeled anomalies. 
Results are reported in \Autoref{table:use-case-eval}. For all metrics except Specificity, the MC-integrated model does better. Interestingly, the base model has better Specificity while the MC model has better Sensitivity - this indicates the MC model is better at identifying the anomalies (the true negatives) and the base is better at identifying the normal traces (the true positives). Example anomaly predictions 
% between the two models 
% for a trace with an obvious anomaly is 
are shown in \Autoref{fig:AD-depiction} in \ref{sec:appdx-ad}.

\begin{table}[t]
\centering
\caption{Use Case Performance Summary. The arrow indicates desired result direction and bold values indicate the best performing model.}\label{table:use-case-eval}
\resizebox{\textwidth}{!}{%
\begin{tabular}{c|c|c|cccc|cccc}\toprule
\multirow{2}{*}{\textbf{Dataset}} & \multirow{2}{*}{\textbf{Model}} & \textbf{Forecasting} & \multicolumn{4}{c|}{\textbf{Anomaly Detection}} & \multicolumn{4}{c}{\textbf{Clustering}} \\ %\hline
& & RMSE ($\downarrow$) & Accuracy ($\uparrow$) & F1 ($\uparrow$) & Sensitivity ($\uparrow$) & Specificity ($\uparrow$) & C-Index ($\downarrow$) & SSE ($\downarrow$) & Silhouette ($\uparrow$) & Cali\'nski Harabasz ($\uparrow$) \\ \toprule
\multirow{2}{*}{Glucose} & Base & $0.22\pm0.05$ & $0.83\pm0.02$ & $0.80\pm0.03$ & $0.67\pm0.02$ & $\mathbf{1.0\pm0.0}$ & $0.086\pm1e{-6}$ & $477.0\pm1e{-3}$ & $0.318\pm1e{-2}$ & $490.95\pm6e{-2}$ \\ %\midrule
& MC & $\mathbf{0.18\pm0.07}$ & $\mathbf{0.92\pm0.02}$ & $\mathbf{0.92\pm0.02}$ & $\mathbf{1.0\pm0.0}$ & $0.85\pm0.03$  & $\mathbf{0.085\pm3e{-5}}$ & $\mathbf{473.79\pm5e{-3}}$ &  $\mathbf{0.332\pm2e{-3}}$ & $\mathbf{496.24\pm5e{-4}}$ \\ \midrule
\multirow{2}{*}{ECG} & Base & $0.30\pm0.14$ & $0.72\pm0.10$ & $0.74\pm0.11$ & $0.59\pm0.13$ & $0.70\pm0.09$ & $0.041\pm6e{-2}$ & $821.10\pm2e{-3}$ & $0.426\pm3e{-3}$ & $970.82\pm4e{-3}$ \\
& MC & $\mathbf{0.17\pm0.12}$ & $\mathbf{0.80\pm0.11}$ & $\mathbf{0.75\pm0.09}$ & $\mathbf{0.83\pm0.09}$ & $0.70\pm0.12$ & $\mathbf{0.040\pm3e{-2}}$ & $\mathbf{820.46\pm1e{-2}}$ & $\mathbf{0.428\pm4e{-3}}$ & $\mathbf{972.27\pm2e{-4}}$ \\ \midrule

\multirow{2}{*}{Resp} & Base & $0.32\pm0.11$ & $0.65\pm0.12$ & $0.73\pm0.13$ &  $0.31\pm0.14$ & $\mathbf{0.91\pm0.15}$ & $0.075\pm2e{-3}$ & $476.6\pm3e{-4}$ & $0.430\pm3e{-2}$ & $1006.12\pm5e{-2}$ \\
& MC & $\mathbf{0.26\pm0.14}$ & $\mathbf{0.71\pm0.12}$ & $\mathbf{0.74\pm0.11}$ & $\mathbf{0.65\pm0.12}$ & $0.86\pm0.14$ & $\mathbf{0.073\pm5e{-3}}$ & $\mathbf{476.5\pm2e{-3}}$ & $\mathbf{0.432\pm4e{-2}}$ & $\mathbf{1007.65\pm3e{-4}}$ \\ \bottomrule
\end{tabular}
}
\end{table}

\boldsection{Clustering}
We implement a simple K-Means clustering algorithm.
In the base model we compute distances between the cluster centroids and other trace data points using Discrete Time Warping (DTW) \citep{sakoe1978dynamic}.
In the MC-integrated version we compute the distances as DTW + MC. We set the number of clusters $k=3$, and $\tau=48$. Causal discovery and clustering models were trained for 20 epochs. 
We compute a set of clustering evaluation metrics including C-Index \citep{hubert1976general}, Sum of Squared Error (SSE) \citep{macqueen1967some}, 
% Davies Bouldin score \citep{davies1979cluster},
Silhouette score \citep{rousseeuw1987silhouettes}, and Cali\'nski Harabasz score \citep{calinski1974dendrite}, reported in \Autoref{table:use-case-eval}. Across all metrics the MC version does better, indicating adding an element of causality may help  clustering.

%% file: 5.conclusion.tex
\section{Conclusion \& Limitations}\label{sec:conc}
In this paper we presented \MD, the first causal discovery framework to infer Motif Causality amongst time series motifs. 
By providing a new method to learn and quantify relationships amongst motifs, \MD may facilitate the development of advanced, high performing 
% models and applications 
technologies for event-based time series.
As shown by the scalability experiments, for very large $\mathcal{M}$ and $n$ (e.g., $n\geq10000$) the runtime can take several hours. There are many opportunities in the training framework to further optimize the runtime. 
% For example, currently to compute the MC values between motifs in a batch, each edge MC weight is computed sequentially. 
For example, MC is currently computed between each edge in a batch sequentially; 
using sampling or parallelization would significantly speed up the training time.
Additionally, a challenge of this work is that there is no known causal structure available, so it was not possible to evaluate the learned motif causal graphs against some ground truth.
However, as evidenced by the use case evaluation, using a relatively simple integration of MC with naive base models resulted in significant performance improvements for all three use cases, providing some evidence 
% that the MC graphs are learning something useful, and supports the 
about the potential generalizability and applicability of MC for many real world tasks.

%% file: 6.appendix.tex
\newpage
\appendix
\section{Technical Appendices and Supplementary Material}\label{sec:appdx}

\subsection{Additional Introduction}\label{sec:appdx-intro}
\begin{figure}[h]
    \centering
    \includegraphics[width=\linewidth]{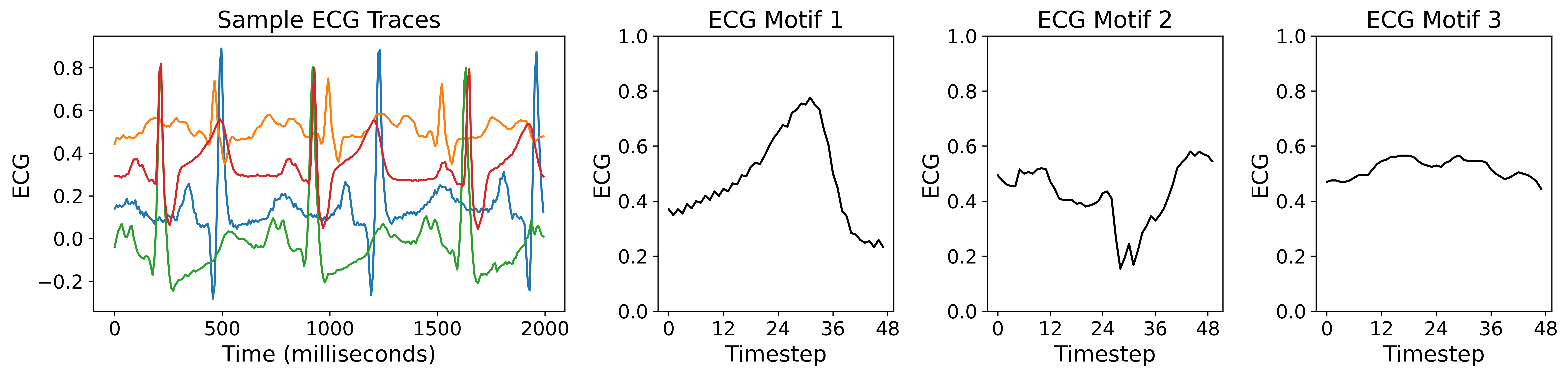}
    \caption{Real ECG Traces and Sample Motifs for $\tau=48$.}
    \label{fig:sample-ecg}
\end{figure}

\begin{figure}[h]
    \centering
    \includegraphics[width=\linewidth]{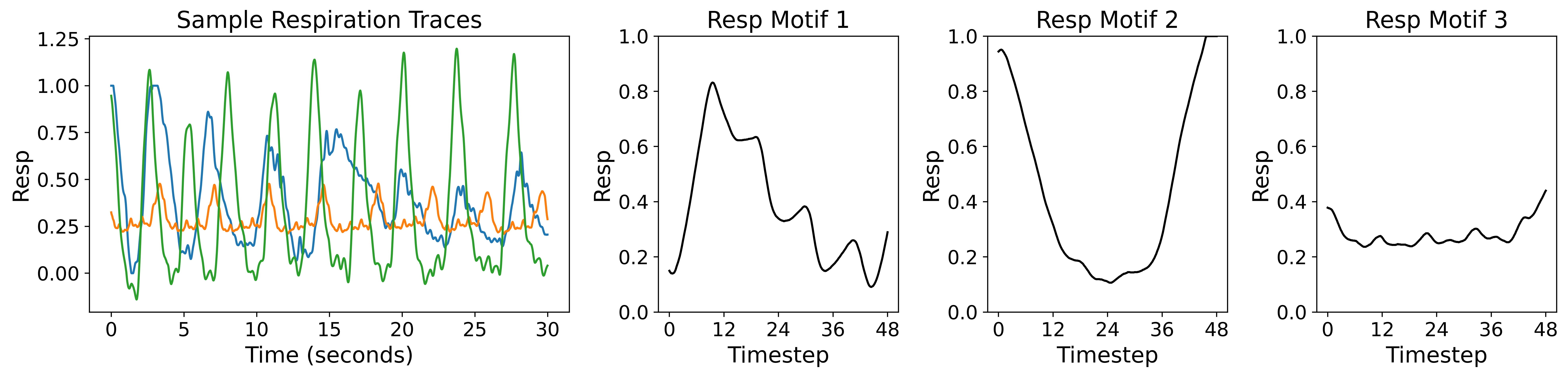}
    \caption{Real Respiration (Resp) Traces and Sample Motifs for $\tau=48$.}
    \label{fig:sample-resp}
\end{figure}

\boldsection{Sample Traces} Sample real traces and motifs from glucose are available in \Autoref{fig:sample-motifs}. Sample traces and motifs for ECG are shown in \Autoref{fig:sample-ecg} and for respiration (resp) are shown in \Autoref{fig:sample-resp}.

\subsection{Additional Methodology}\label{sec:appdx-methods}

\boldsection{Extracting Motifs}
We assume there is a consistent, conclusive way to pull motifs from the traces. Motif discovery is outside the scope of this work; discussion about motif extraction methods and artifacts is outside the scope of the paper since the problem of how to extract motifs has been widely solved (e.g., there are many nice methods to pull out motifs in an intelligent manner, given some criteria, and specific to different disciplines, such as ~\cite{chinpattanakarn2024framework, schafer2022motiflets, ye2009time}). Our framework is amenable to any motif extraction method, and, as evidenced by our results, works even when using simple or heuristic motif extraction methods, such as chopping. 
% We refer the interested reader to other works focused on this problem \citep{chinpattanakarn2024framework, schafer2022motiflets, ye2009time}. 

\begin{figure}[ht]
     \centering
      \subfigure[Large Uncertainty Reduction]{
         \centering
         \label{fig:big-red}
         \includegraphics[width=0.49\textwidth]{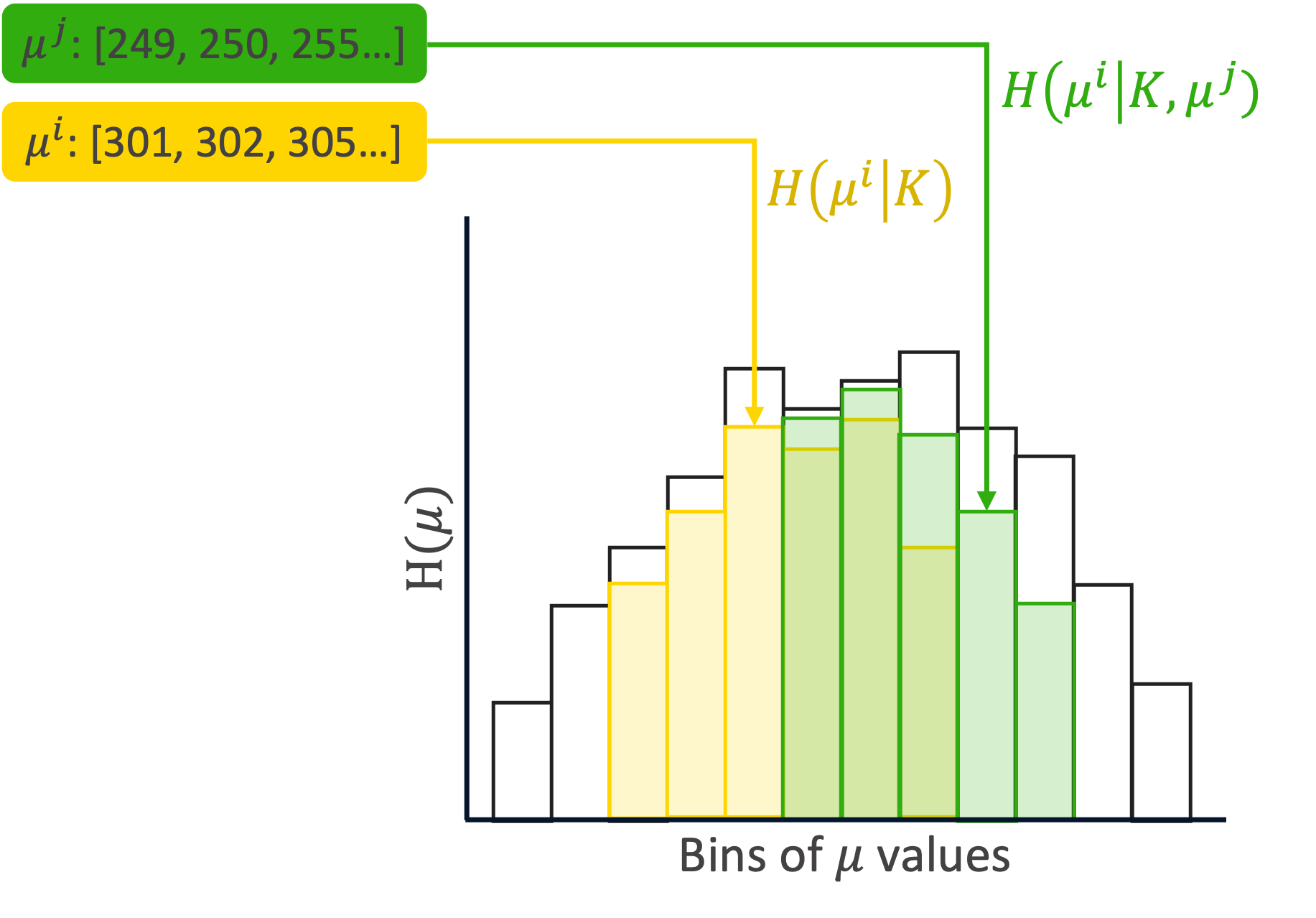}}
        \hfill
     \subfigure[Small Uncertainty Reduction]{
         \centering
         \label{fig:no-red}
         \includegraphics[width=0.49\textwidth]{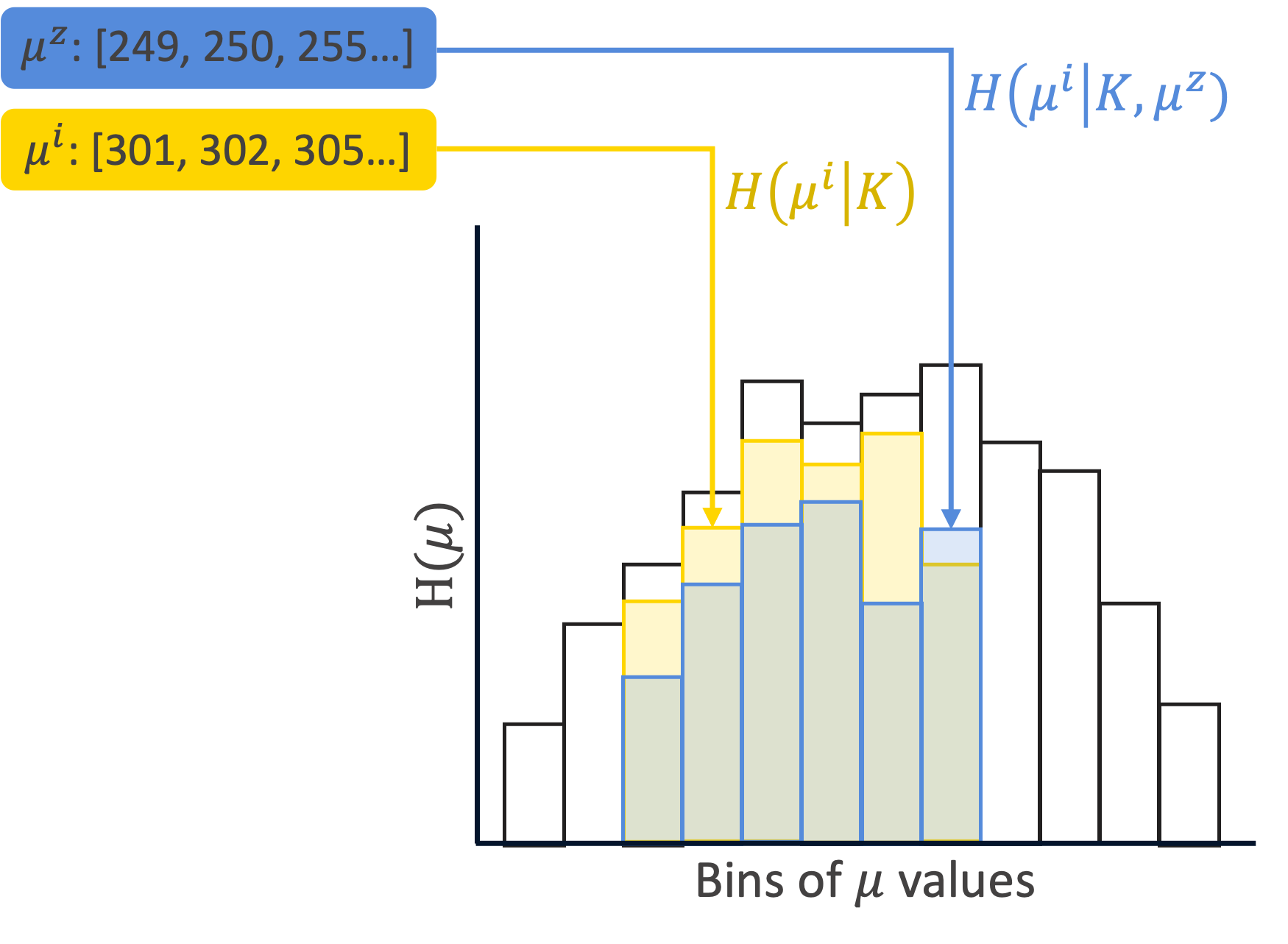}}
    \caption{Depiction of Conditional Entropy $H(\mu)$ computation using the histogram method: when computing MC for $\mu^i$, \subref{fig:big-red} $\mu^j$ results in a large uncertainty reduction; \subref{fig:no-red} $\mu^z$ has a small reduction.}
    \label{fig:TE-hist}
\end{figure}

\boldsection{MC Computation in Link Predictor} To implement the MC computation, we adapt existing Transfer Entropy libraries \citep{behrendt2019rtransferentropy} and compute the conditional entropy function (i.e., $H(\mu)$, Shannon's or R\'enyi entropy) using the histogram method. A simplified depiction is shown in \Autoref{fig:TE-hist}. Essentially, motif time series values are binned into a histogram.
% bins.
% are binned based on their values into histogram bins. 
% From there, 
Distributions between motifs can be compared to determine how much uncertainty about future predictions of the motif is reduced in the distribution. For example, in \ref{fig:big-red} when computing MC for $\mu^i$, $\mu^j$ covers a larger distribution, resulting in a large reduction in uncertainty and higher motif causality, whereas \ref{fig:no-red} $\mu^z$ covers hardly any new distribution compared to the status quo ($H(\mu^i|\mathcal{K})$), resulting in a small reduction in uncertainty and low causality for $\mu^z$.

\begin{figure}[ht]
    \centering
    \includegraphics[width=0.4\textwidth]{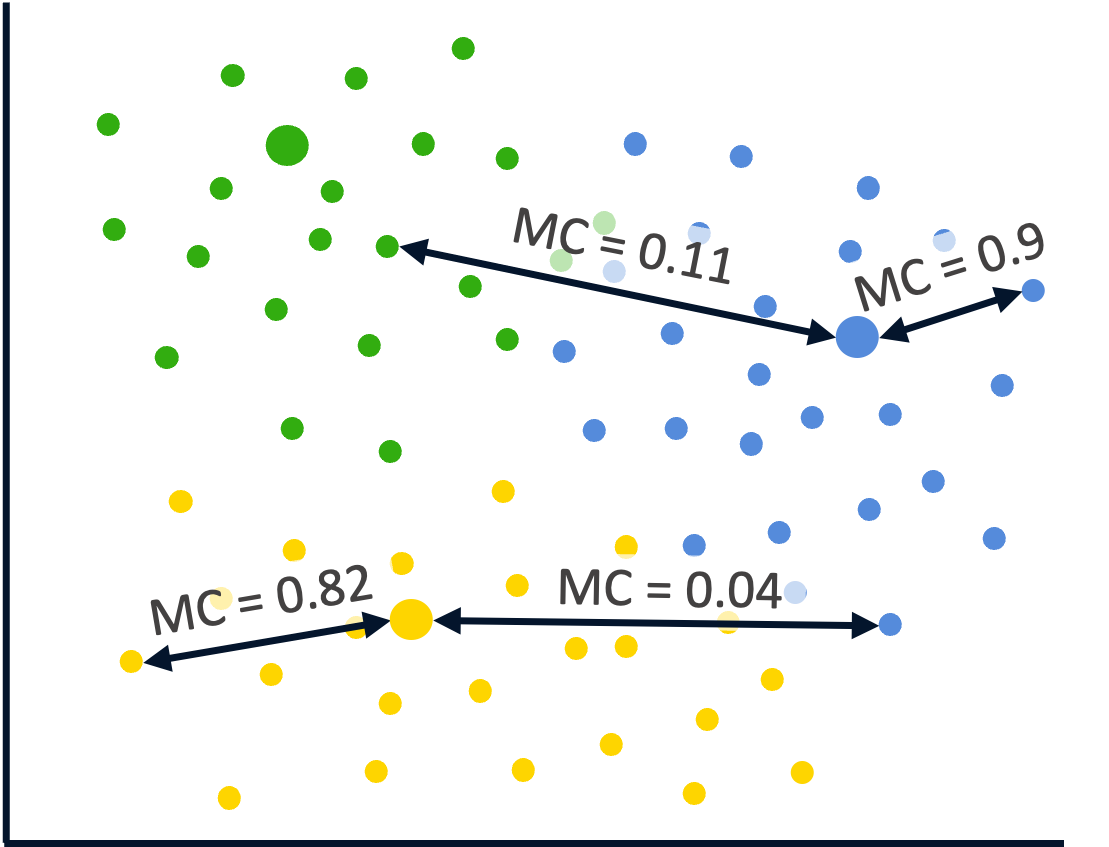}
    \caption[width=0.4\textwidth]{Depiction of Clustering Model integrated with MC.}
    \label{fig:clustering}
\end{figure}

\boldsection{Clustering Use Case} 
To integrate Motif Causality with a basic clustering algorithm, 
% when computing the distance metric we also compute the Motif causality values 
we use the motif causality values as an additional distance metric. 
The intuition is to add an element of \textit{causality} to the clustering, such that as the algorithm learns, MC values within each cluster will be minimized and similar data points within the cluster should be \textit{motif causal} of each other. An example is shown in \Autoref{fig:clustering}: the MC value between the blue centroid and the blue data point to the right is high with 0.9, whereas the MC between the blue centroid and the green data point belonging to a different cluster is low at 0.11.

\subsection{Additional Evaluation}\label{sec:appdx-eval}

As mentioned in \Autoref{sec:rel-work}, there is no prior work on quantifying causality amongst motifs, and we cannot compare with existing Granger causal techniques because their framework set-up or model assumptions do not hold. Specifically, they have one or more of the following issues: 
\begin{itemize}
    \item Compose relations between only two time series, and cannot find relationships amongst a \textit{set} of motifs or traces \cite{gong2023causal, amornbunchornvej2021variable}.
    \item Formulate causality based on various statistical properties amongst \textit{variables}, such as variable-based correlation and density-, classification- and prediction-based error measures computed between multivariate features  \cite{bonetti2024causal, amornbunchornvej2021variable, irribarra2024multi}. These assumptions do not hold for \textit{univariate} time series motifs.
    \item Compose causal relationships over time series using repeated statistical measures or temporal dynamics, e.g., things like learning implicit repeated patterns, or computing sufficiency or faithfulness hypotheses over lags across the full time series \cite{lamp2024glucosynth, pan2024effcause, tank2021neural, lowe2022amortized, najafi2023entropy, assaad2021mixed, sun2015causal}. These methods do not work for motifs because they are short time series (so there would be no repeated patterns, nor will statistical hypotheses like sufficiency or faithfulness hold since they are not evaluating multiple time lags repeated over a trace).
    \item Require labels or a known underlying causal structure to guide model training \cite{gong2023causal, bonetti2024causal, najafi2023entropy}, which is not available for our data and many similar event-based data streams.
\end{itemize}

Therefore, for our evaluation in \Autoref{sec:eval}, we evaluate our causal discovery framework in terms of scalability and performance for 3 downstream use cases: forecasting, anomaly detection and clustering. 

\subsubsection{Dataset Details}
An overview of characteristics for each dataset is shown in \Autoref{table:data-chars}. The glucose traces are sets of single-day glucose traces randomly sampled across each month from January to December 2022, collected from Dexcom's G6 Continuous Glucose Monitors (CGMs)~\cite{akturk2021real}. Data was recorded every 5 minutes, resulting in a total of 288 timepoints per trace. Each trace was aligned temporally from 00:00 to 23:59. The ECG and Respiration (resp) traces come from  MIMIC-BP~\cite{dvn_2023}, a curated dataset collecting common biomedical signals. We use data from a single segment. There are a total of 1523 patients, with data collected for a total of 3750 timepoints, sampled at frequencies of 1 millisecond for ECG and 1 second for Resp.

\begin{table}[h]
\centering
\caption{Dataset Characteristics.}\label{table:data-chars}
% \resizebox{\textwidth}{!}{%
\begin{tabular}{c|c|c|c}\toprule
\textbf{Dataset} & \textbf{\# Patients} & \textbf{Total \# Timepoints} & \textbf{Sampling Frequency} \\ \toprule
Glucose~\cite{akturk2021real} & 300000 & 288 & 5 minutes \\
ECG~\cite{dvn_2023} & 1523 & 3750 & 1 millisecond \\
Resp~\cite{dvn_2023} & 1523 & 3750 & 1 second \\ \bottomrule
\end{tabular}
% }
\end{table}

\subsubsection{Additional Scalability Results}\label{sec:appdx-scale}
\begin{figure}[ht]
     \centering
      \subfigure[$n$ vs. Training Time]{
         \centering
         \label{fig:scale-traceecg}
         \includegraphics[width=0.32\textwidth]{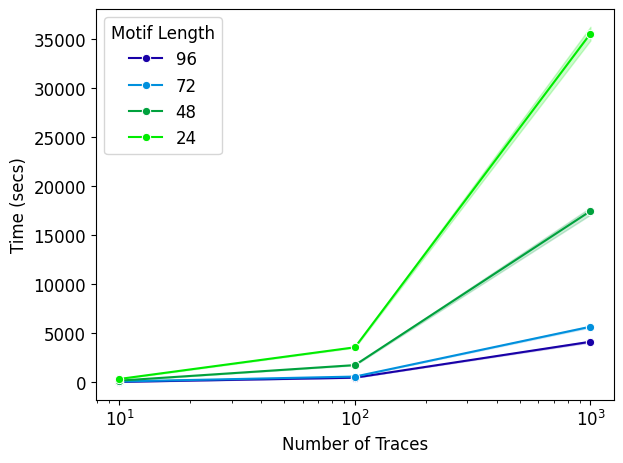}}
        \hfill
     \subfigure[$\tau$ vs. Training Time]{
         \centering
         \label{fig:scale-motifecg}
         \includegraphics[width=0.32\textwidth]{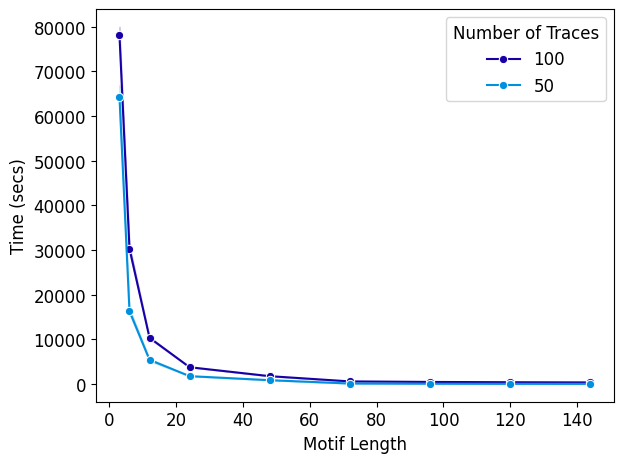}}
         \hfill
     \subfigure[$|\mathcal{M}|$ vs. Training Time]{
         \centering
         \label{fig:scale-motifsetecg}
         \includegraphics[width=0.32\textwidth]{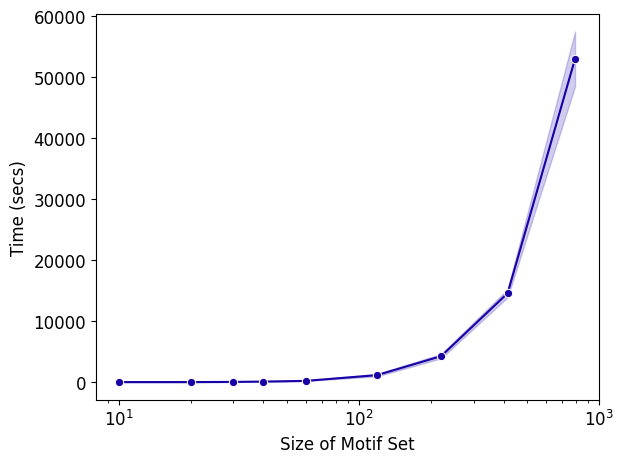}}
    \caption{Scalability for ECG traces varying \subref{fig:scale-traceecg} \# of traces ($n$), \subref{fig:scale-motifecg} motif lengths ($\tau$)  and \subref{fig:scale-motifsetecg} motif set sizes ($|\mathcal{M}|$). }
    \label{fig:scalability-ecg}
\end{figure}

\begin{figure}[ht]
     \centering
      \subfigure[$n$ vs. Training Time]{
         \centering
         \label{fig:scale-traceresp}
         \includegraphics[width=0.32\textwidth]{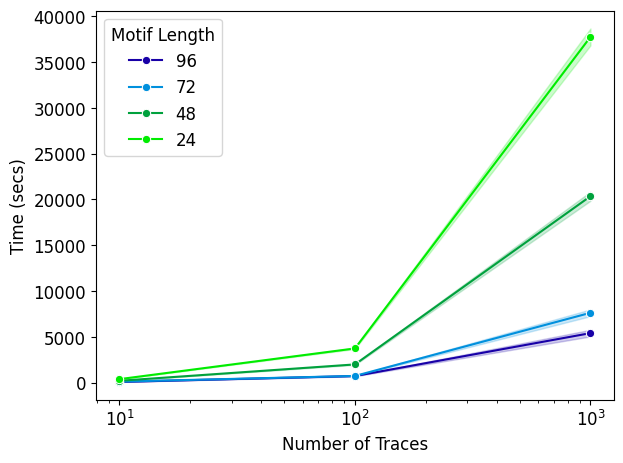}}
        \hfill
     \subfigure[$\tau$ vs. Training Time]{
         \centering
         \label{fig:scale-motifresp}
         \includegraphics[width=0.32\textwidth]{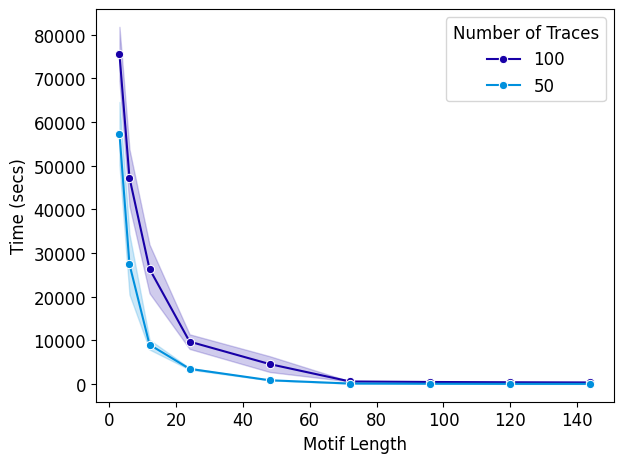}}
         \hfill
     \subfigure[$|\mathcal{M}|$ vs. Training Time]{
         \centering
         \label{fig:scale-motifsetresp}
         \includegraphics[width=0.32\textwidth]{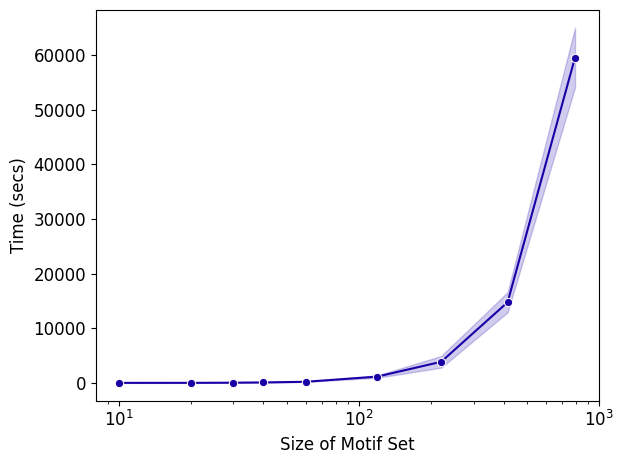}}
    \caption{Scalability for respiration traces varying \subref{fig:scale-traceresp} \# of traces ($n$), \subref{fig:scale-motifresp} motif lengths ($\tau$)  and \subref{fig:scale-motifsetresp} motif set sizes ($|\mathcal{M}|$).}
    \label{fig:scalability-resp}
\end{figure}

Due to space constraints, we report the results for the MIMIC-BP dataset streams of ECG and Respiration (Resp) in \Autoref{fig:scalability-ecg} and \Autoref{fig:scalability-resp} respectively here. As found with the glucose traces, training time increases for larger $n$ and $|\mathcal{M}|$.
Time also reduces as the $\tau$ increases, which may in part be because larger $\tau$s mean there are less total motifs (i.e., $|\mathcal{M}|$ is smaller). Overall, training time for the Resp dataset takes the longest, perhaps because the Resp traces are the more variable, resulting in more variable (i.e., more noisy) motifs.

\subsubsection{Additional Use Case Results}\label{sec:appdx-ad}
\begin{figure}[ht]
     \centering
     \vspace{-0.25cm}
      \subfigure[Base Model]{
         \centering
         \label{fig:AD-base}
         \includegraphics[width=0.42\textwidth]{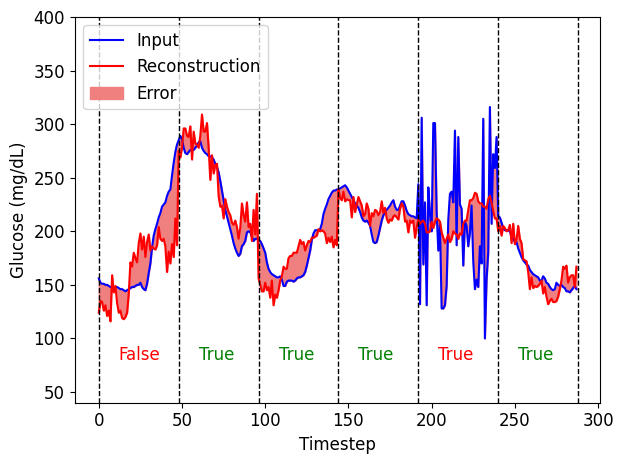}}
        % \hfill
     \subfigure[Base Model + MC]{
         \centering
         \label{fig:AD-MC}
         \includegraphics[width=0.42\textwidth]{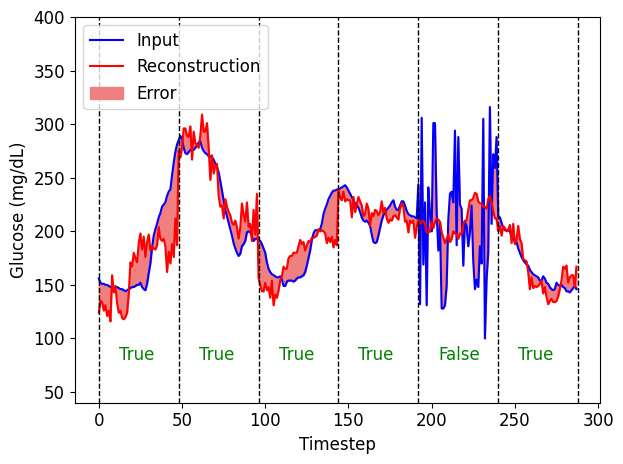}}
    \vspace{-0.25cm}
    \caption{Anomaly Detection for a trace with an obvious anomaly using the \subref{fig:AD-base} base model and the model integrated with MC \subref{fig:AD-MC}. False means an anomaly was predicted and the color corresponds to correctness of the prediction: green is a correct prediction, red is an incorrect one.}
    \label{fig:AD-depiction}
\end{figure}

\boldsection{Anomaly Detection} Example anomaly predictions 
% between the two models 
% for a trace with an obvious anomaly is 
are shown in \Autoref{fig:AD-depiction}. Each graph plots the original input trace (in blue) and the reconstructed trace (in red, trace fed through the encoder + decoder) with the error between the two shaded in red. The window segments (dashed black lines) correspond to the motif and prediction window size and the model predictions are annotated in each window, colored by the correctness of the prediction. Green indicates a correct prediction, red indicates an incorrect one.

\subsubsection{Example Motif Causality Between Motifs }\label{sec:apdx-mcfigs}
\begin{figure}
     \centering
      \subfigure[$\tau=6$]{
         \centering
         \label{fig:MC-example-6}
         \includegraphics[width=0.45\textwidth]{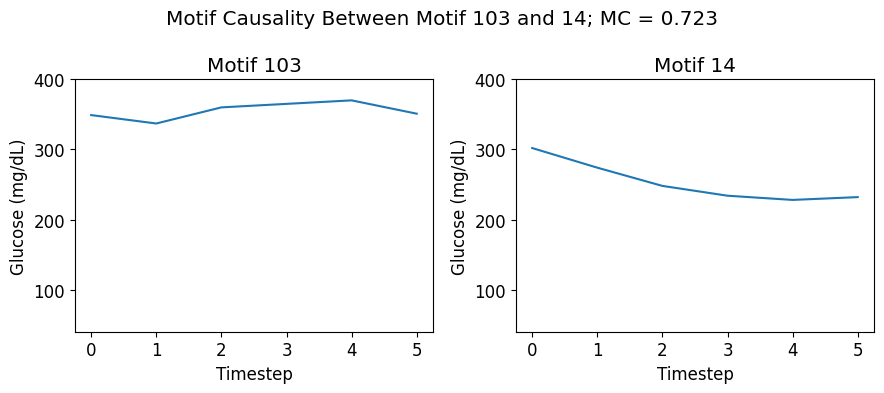}}
        \hfill
     \subfigure[$\tau=12$]{
         \centering
         \label{fig:MC-example-12}
         \includegraphics[width=0.45\textwidth]{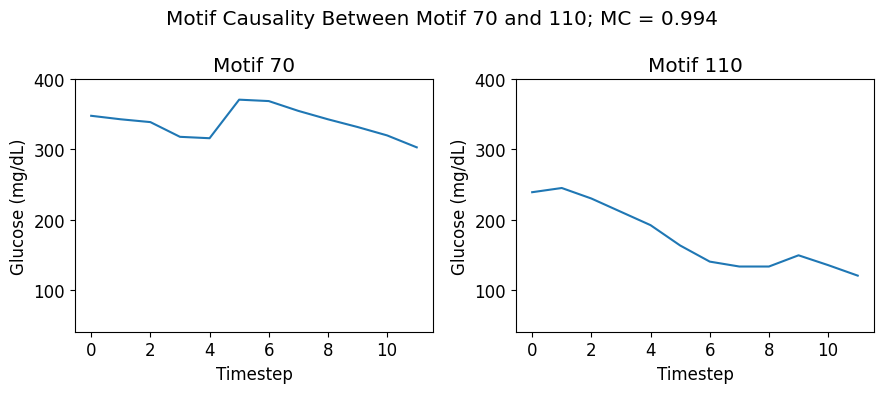}}
        \hfill
    \subfigure[$\tau=48$]{
         \centering
         \label{fig:MC-example-48}
         \includegraphics[width=0.45\textwidth]{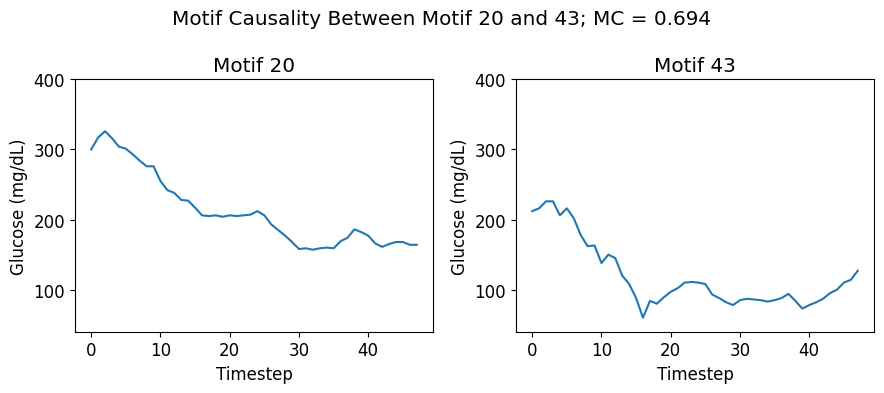}}
        \hfill
    % \subfigure[$\tau=96$]{
    %      \centering
    %      \label{fig:MC-example-6}
    %      \includegraphics[width=0.8\textwidth]{Figures/Motif_Preds_nseq10_tau96_epochs30_strongTE1.png}}
    %     \hfill
    % \subfigure[$\tau=144$ Low]{
    %      \centering
    %      \label{fig:MC-example-6}
    %      \includegraphics[width=0.8\textwidth]{Figures/Motif_Preds_nseq10_tau144_epochs20_lowTE1.png}}
    %     \hfill
    \subfigure[$\tau=144$]{
         \centering
         \label{fig:MC-example-144}
         \includegraphics[width=0.45\textwidth]{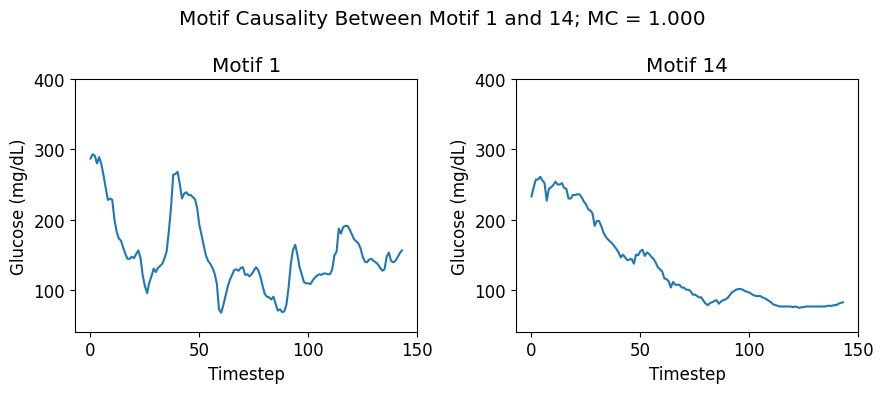}}
        \hfill
    \caption{Example High Motif Causality values between different motif sizes for Glucose.}
    \label{fig:MC-examples-strong}
\end{figure}

\begin{figure}
     \centering
      \subfigure[$\tau=6$]{
         \centering
         \label{fig:MC-example-6low}
         \includegraphics[width=0.45\textwidth]{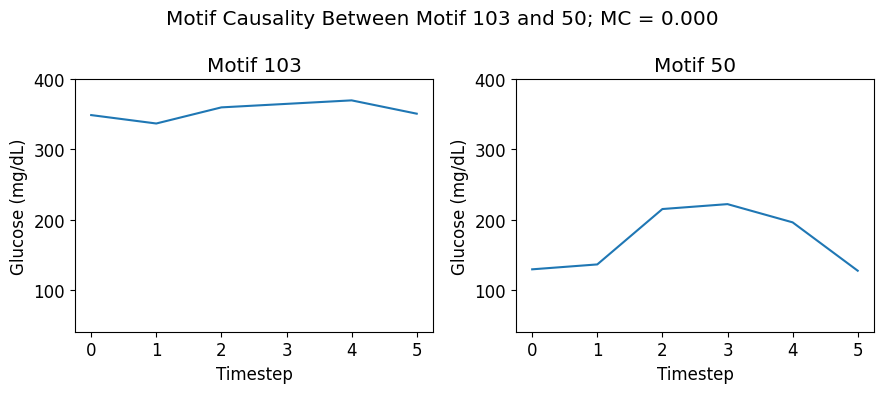}}
        \hfill
     \subfigure[$\tau=12$]{
         \centering
         \label{fig:MC-example-12low}
         \includegraphics[width=0.45\textwidth]{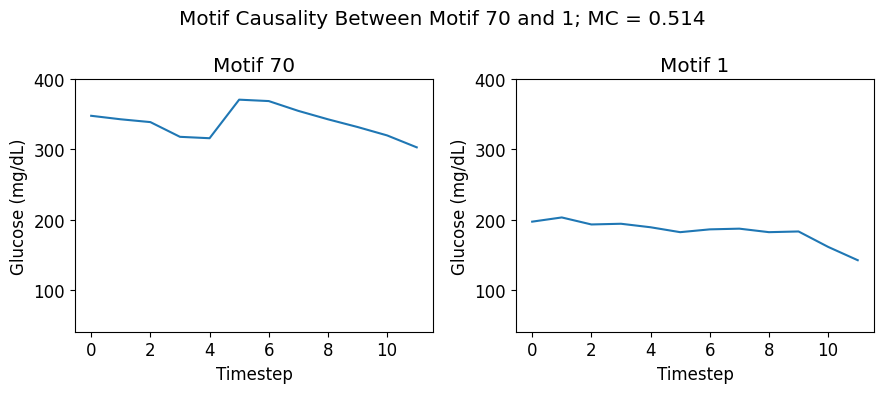}}
        \hfill
    \subfigure[$\tau=96$]{
         \centering
         \label{fig:MC-example-96low}
         \includegraphics[width=0.45\textwidth]{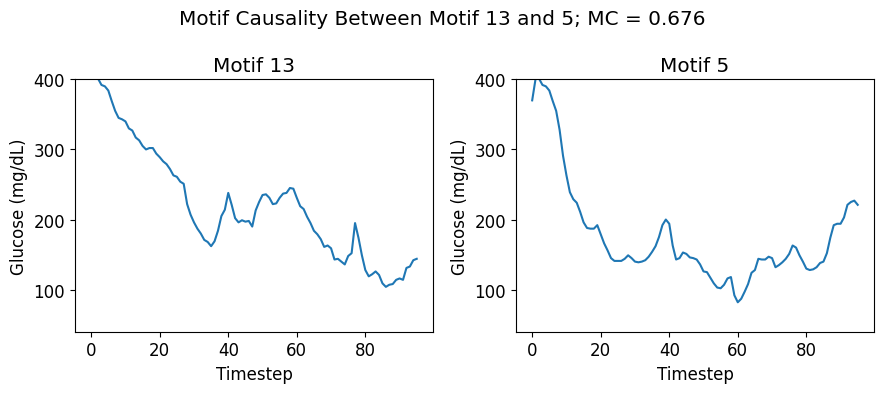}}
        \hfill
    % \subfigure[$\tau=96$]{
    %      \centering
    %      \label{fig:MC-example-6}
    %      \includegraphics[width=0.8\textwidth]{Figures/Motif_Preds_nseq10_tau96_epochs30_strongTE1.png}}
    %     \hfill
    % \subfigure[$\tau=144$ Low]{
    %      \centering
    %      \label{fig:MC-example-6}
    %      \includegraphics[width=0.8\textwidth]{Figures/Motif_Preds_nseq10_tau144_epochs20_lowTE1.png}}
    %     \hfill
    \subfigure[$\tau=144$]{
         \centering
         \label{fig:MC-example-144low}
         \includegraphics[width=0.45\textwidth]{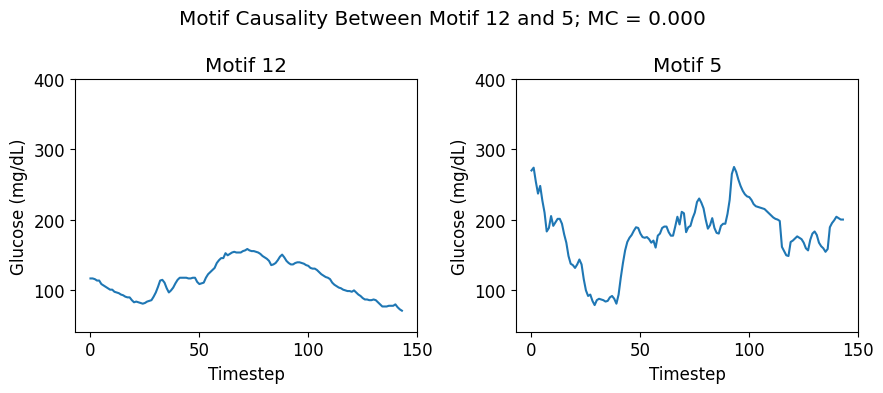}}
        \hfill
    \caption{Example Low Motif Causality values between different motif sizes for Glucose.}
    \label{fig:MC-examples-low}
\end{figure}

Learned motif causality values between sample motifs for glucose are illustrated in \Autoref{fig:MC-examples-strong} and \Autoref{fig:MC-examples-low}.

\subsection{Additional Conclusion}\label{sec:apdx-conc}
\boldsection{Broader Impacts \& Safeguards} The core contribution of this project is the development of \MD, the first causal discovery framework to infer Motif Causality amongst time series motifs. In terms of broader impacts, \MD may facilitate the development of advanced, high performing 
% models and applications 
technologies for event-based time series by providing a new method to learn and quantify relationships amongst motifs. Since this project presents a foundational research framework that can be applied in many diverse settings and scenarios, discussion of deployments is outside the scope of this work. However, the authors note that any deployment using this framework may need to integrate the necessary measures and safeguards to reduce the risk of possible societal harms (particularly for applications in medical settings), such as data release and consent frameworks, and strict privacy and security measures.

%% file: neurips_2025.bbl
\begin{thebibliography}{45}
\providecommand{\natexlab}[1]{#1}
\providecommand{\url}[1]{\texttt{#1}}
\expandafter\ifx\csname urlstyle\endcsname\relax
  \providecommand{\doi}[1]{doi: #1}\else
  \providecommand{\doi}{doi: \begingroup \urlstyle{rm}\Url}\fi

\bibitem[Niu et~al.(2024)Niu, Gao, Song, and Li]{niu2024comprehensive}
Wenjin Niu, Zijun Gao, Liyan Song, and Lingbo Li.
\newblock Comprehensive review and empirical evaluation of causal discovery algorithms for numerical data.
\newblock \emph{arXiv preprint arXiv:2407.13054}, 2024.

\bibitem[Granger(1969)]{granger1969investigating}
Clive~WJ Granger.
\newblock Investigating causal relations by econometric models and cross-spectral methods.
\newblock \emph{Econometrica: journal of the Econometric Society}, pages 424--438, 1969.

\bibitem[Schreiber(2000)]{schreiber2000measuring}
Thomas Schreiber.
\newblock Measuring information transfer.
\newblock \emph{Physical review letters}, 85\penalty0 (2):\penalty0 461, 2000.

\bibitem[Gong et~al.(2023)Gong, Yao, Zhang, Li, and Bi]{gong2023causal}
Chang Gong, Di~Yao, Chuzhe Zhang, Wenbin Li, and Jingping Bi.
\newblock Causal discovery from temporal data: An overview and new perspectives.
\newblock \emph{arXiv preprint arXiv:2303.10112}, 2023.

\bibitem[Assaad et~al.(2022)Assaad, Devijver, and Gaussier]{assaad2022survey}
Charles~K Assaad, Emilie Devijver, and Eric Gaussier.
\newblock Survey and evaluation of causal discovery methods for time series.
\newblock \emph{Journal of Artificial Intelligence Research}, 73:\penalty0 767--819, 2022.

\bibitem[Liu et~al.(2021)Liu, Guarrasi, and Sar{\i}y{\"u}ce]{liu2021temporal}
Penghang Liu, Valerio Guarrasi, and Ahmet~Erdem Sar{\i}y{\"u}ce.
\newblock Temporal network motifs: Models, limitations, evaluation.
\newblock \emph{IEEE Transactions on Knowledge and Data Engineering}, 35\penalty0 (1):\penalty0 945--957, 2021.

\bibitem[Kovanen et~al.(2011)Kovanen, Karsai, Kaski, Kert{\'e}sz, and Saram{\"a}ki]{kovanen2011temporal}
Lauri Kovanen, M{\'a}rton Karsai, Kimmo Kaski, J{\'a}nos Kert{\'e}sz, and Jari Saram{\"a}ki.
\newblock Temporal motifs in time-dependent networks.
\newblock \emph{Journal of Statistical Mechanics: Theory and Experiment}, 2011\penalty0 (11):\penalty0 P11005, 2011.

\bibitem[Chen and Ying(2024)]{chen2024tempme}
Jialin Chen and Rex Ying.
\newblock Tempme: Towards the explainability of temporal graph neural networks via motif discovery.
\newblock \emph{Advances in Neural Information Processing Systems}, 36, 2024.

\bibitem[Chen et~al.(2023{\natexlab{a}})Chen, Cai, Fang, Wu, Li, and Hao]{chen2023motif}
Xuexin Chen, Ruichu Cai, Yuan Fang, Min Wu, Zijian Li, and Zhifeng Hao.
\newblock Motif graph neural network.
\newblock \emph{IEEE Transactions on Neural Networks and Learning Systems}, 2023{\natexlab{a}}.

\bibitem[Jin et~al.(2022)Jin, Li, and Pan]{jin2022neural}
Ming Jin, Yuan-Fang Li, and Shirui Pan.
\newblock Neural temporal walks: Motif-aware representation learning on continuous-time dynamic graphs.
\newblock \emph{Advances in Neural Information Processing Systems}, 35:\penalty0 19874--19886, 2022.

\bibitem[Chinpattanakarn and Amornbunchornvej(2024)]{chinpattanakarn2024framework}
Naaek Chinpattanakarn and Chainarong Amornbunchornvej.
\newblock Framework for variable-lag motif following relation inference in time series using matrix profile analysis.
\newblock \emph{arXiv preprint arXiv:2401.02860}, 2024.

\bibitem[Lamp et~al.(2024)Lamp, Derdzinski, Hannemann, Van~der Linden, Feng, Wang, and Evans]{lamp2024glucosynth}
Josephine Lamp, Mark Derdzinski, Christopher Hannemann, Joost Van~der Linden, Lu~Feng, Tianhao Wang, and David Evans.
\newblock Glucosynth: Generating differentially-private synthetic glucose traces.
\newblock \emph{Advances in Neural Information Processing Systems}, 36, 2024.

\bibitem[Hasan et~al.(2023)Hasan, Hossain, and Gani]{hasan2023survey}
Uzma Hasan, Emam Hossain, and Md~Osman Gani.
\newblock A survey on causal discovery methods for iid and time series data.
\newblock \emph{arXiv preprint arXiv:2303.15027}, 2023.

\bibitem[Shojaie and Fox(2022)]{shojaie2022granger}
Ali Shojaie and Emily~B Fox.
\newblock Granger causality: A review and recent advances.
\newblock \emph{Annual Review of Statistics and Its Application}, 9\penalty0 (1):\penalty0 289--319, 2022.

\bibitem[Pan et~al.(2024)Pan, Zhang, Jiang, Ma, and Wang]{pan2024effcause}
Yicheng Pan, Yifan Zhang, Xinrui Jiang, Meng Ma, and Ping Wang.
\newblock Effcause: Discover dynamic causal relationships efficiently from time-series.
\newblock \emph{ACM Transactions on Knowledge Discovery from Data}, 18\penalty0 (5):\penalty0 1--21, 2024.

\bibitem[Tank et~al.(2021)Tank, Covert, Foti, Shojaie, and Fox]{tank2021neural}
Alex Tank, Ian Covert, Nicholas Foti, Ali Shojaie, and Emily~B Fox.
\newblock Neural granger causality.
\newblock \emph{IEEE Transactions on Pattern Analysis and Machine Intelligence}, 44\penalty0 (8):\penalty0 4267--4279, 2021.

\bibitem[L{\"o}we et~al.(2022)L{\"o}we, Madras, Zemel, and Welling]{lowe2022amortized}
Sindy L{\"o}we, David Madras, Richard Zemel, and Max Welling.
\newblock Amortized causal discovery: Learning to infer causal graphs from time-series data.
\newblock In \emph{Conference on Causal Learning and Reasoning}, pages 509--525. PMLR, 2022.

\bibitem[Bonetti et~al.(2024)Bonetti, Metelli, and Restelli]{bonetti2024causal}
Paolo Bonetti, Alberto~Maria Metelli, and Marcello Restelli.
\newblock Causal feature selection via transfer entropy.
\newblock In \emph{2024 International Joint Conference on Neural Networks (IJCNN)}, pages 1--10. IEEE, 2024.

\bibitem[Najafi et~al.(2023)Najafi, Parsaeefard, and Leon-Garcia]{najafi2023entropy}
Bahareh Najafi, Saeedeh Parsaeefard, and Alberto Leon-Garcia.
\newblock Entropy-aware time-varying graph neural networks with generalized temporal hawkes process: Dynamic link prediction in the presence of node addition and deletion.
\newblock \emph{Machine Learning and Knowledge Extraction}, 5\penalty0 (4):\penalty0 1359--1381, 2023.

\bibitem[Ansari et~al.(2024)Ansari, Stella, Turkmen, Zhang, Mercado, Shen, Shchur, Rangapuram, Arango, Kapoor, et~al.]{ansari2024chronos}
Abdul~Fatir Ansari, Lorenzo Stella, Caner Turkmen, Xiyuan Zhang, Pedro Mercado, Huibin Shen, Oleksandr Shchur, Syama~Sundar Rangapuram, Sebastian~Pineda Arango, Shubham Kapoor, et~al.
\newblock Chronos: Learning the language of time series.
\newblock \emph{arXiv preprint arXiv:2403.07815}, 2024.

\bibitem[Chen et~al.(2023{\natexlab{b}})Chen, Chen, Shang, Wu, Zheng, Wen, and Zhang]{chen2023multi}
Ling Chen, Donghui Chen, Zongjiang Shang, Binqing Wu, Cen Zheng, Bo~Wen, and Wei Zhang.
\newblock Multi-scale adaptive graph neural network for multivariate time series forecasting.
\newblock \emph{IEEE Transactions on Knowledge and Data Engineering}, 35\penalty0 (10):\penalty0 10748--10761, 2023{\natexlab{b}}.

\bibitem[Febrinanto et~al.(2023)Febrinanto, Moore, Thapa, Liu, Saikrishna, Ma, and Xia]{febrinanto2023entropy}
Falih~Gozi Febrinanto, Kristen Moore, Chandra Thapa, Mujie Liu, Vidya Saikrishna, Jiangang Ma, and Feng Xia.
\newblock Entropy causal graphs for multivariate time series anomaly detection.
\newblock \emph{arXiv preprint arXiv:2312.09478}, 2023.

\bibitem[Duan et~al.(2022)Duan, Xu, Huang, Feng, and Wang]{duan2022multivariate}
Ziheng Duan, Haoyan Xu, Yida Huang, Jie Feng, and Yueyang Wang.
\newblock Multivariate time series forecasting with transfer entropy graph.
\newblock \emph{Tsinghua Science and Technology}, 28\penalty0 (1):\penalty0 141--149, 2022.

\bibitem[Wu et~al.(2021)Wu, Gu, Wang, Lin, Wang, and Yang]{wu2021event2graph}
Yuhang Wu, Mengting Gu, Lan Wang, Yusan Lin, Fei Wang, and Hao Yang.
\newblock Event2graph: Event-driven bipartite graph for multivariate time-series anomaly detection.
\newblock \emph{arXiv preprint arXiv:2108.06783}, 2021.

\bibitem[Sch{\"a}fer and Leser(2022)]{schafer2022motiflets}
Patrick Sch{\"a}fer and Ulf Leser.
\newblock Motiflets: Simple and accurate detection of motifs in time series.
\newblock \emph{Proceedings of the VLDB Endowment}, 16\penalty0 (4):\penalty0 725--737, 2022.

\bibitem[Ye and Keogh(2009)]{ye2009time}
Lexiang Ye and Eamonn Keogh.
\newblock Time series shapelets: a new primitive for data mining.
\newblock In \emph{ACM SIGKDD International Conference on Knowledge Discovery and Data Mining}, 2009.

\bibitem[Barnett et~al.(2009)Barnett, Barrett, and Seth]{barnett2009granger}
Lionel Barnett, Adam~B Barrett, and Anil~K Seth.
\newblock Granger causality and transfer entropy are equivalent for gaussian variables.
\newblock \emph{Physical review letters}, 103\penalty0 (23):\penalty0 238701, 2009.

\bibitem[Amornbunchornvej et~al.(2021)Amornbunchornvej, Zheleva, and Berger-Wolf]{amornbunchornvej2021variable}
Chainarong Amornbunchornvej, Elena Zheleva, and Tanya Berger-Wolf.
\newblock Variable-lag granger causality and transfer entropy for time series analysis.
\newblock \emph{ACM Transactions on Knowledge Discovery from Data (TKDD)}, 15\penalty0 (4):\penalty0 1--30, 2021.

\bibitem[Irribarra et~al.(2024)Irribarra, Michell, Bermeo, and Kristjanpoller]{irribarra2024multi}
Nicol{\'a}s Irribarra, Kevin Michell, Cristhian Bermeo, and Werner Kristjanpoller.
\newblock A multi-head attention neural network with non-linear correlation approach for time series causal discovery.
\newblock \emph{Applied Soft Computing}, 165:\penalty0 112062, 2024.

\bibitem[Assaad et~al.(2021)Assaad, Devijver, Gaussier, and Ait-Bachir]{assaad2021mixed}
Karim Assaad, Emilie Devijver, Eric Gaussier, and Ali Ait-Bachir.
\newblock A mixed noise and constraint-based approach to causal inference in time series.
\newblock In \emph{Machine Learning and Knowledge Discovery in Databases. Research Track: European Conference, ECML PKDD 2021, Bilbao, Spain, September 13--17, 2021, Proceedings, Part I 21}, pages 453--468. Springer, 2021.

\bibitem[Sun et~al.(2015)Sun, Taylor, and Bollt]{sun2015causal}
Jie Sun, Dane Taylor, and Erik~M Bollt.
\newblock Causal network inference by optimal causation entropy.
\newblock \emph{SIAM Journal on Applied Dynamical Systems}, 14\penalty0 (1):\penalty0 73--106, 2015.

\bibitem[Shannon(1948)]{shannon1948mathematical}
Claude~Elwood Shannon.
\newblock A mathematical theory of communication.
\newblock \emph{The Bell system technical journal}, 27\penalty0 (3):\penalty0 379--423, 1948.

\bibitem[Jizba et~al.(2012)Jizba, Kleinert, and Shefaat]{jizba2012renyi}
Petr Jizba, Hagen Kleinert, and Mohammad Shefaat.
\newblock R{\'e}nyi’s information transfer between financial time series.
\newblock \emph{Physica A: Statistical Mechanics and its Applications}, 391\penalty0 (10):\penalty0 2971--2989, 2012.

\bibitem[Jizba et~al.(2022)Jizba, Lavi{\v{c}}ka, and Tabachov{\'a}]{jizba2022causal}
Petr Jizba, Hynek Lavi{\v{c}}ka, and Zlata Tabachov{\'a}.
\newblock Causal inference in time series in terms of r{\'e}nyi transfer entropy.
\newblock \emph{Entropy}, 24\penalty0 (7):\penalty0 855, 2022.

\bibitem[Rosso et~al.(2001)Rosso, Blanco, Yordanova, Kolev, Figliola, Sch{\"u}rmann, and Ba{\c{s}}ar]{rosso2001wavelet}
Osvaldo~A Rosso, Susana Blanco, Juliana Yordanova, Vasil Kolev, Alejandra Figliola, Martin Sch{\"u}rmann, and Erol Ba{\c{s}}ar.
\newblock Wavelet entropy: a new tool for analysis of short duration brain electrical signals.
\newblock \emph{Journal of neuroscience methods}, 105\penalty0 (1):\penalty0 65--75, 2001.

\bibitem[Bandt and Pompe(2002)]{bandt2002permutation}
Christoph Bandt and Bernd Pompe.
\newblock Permutation entropy: a natural complexity measure for time series.
\newblock \emph{Physical review letters}, 88\penalty0 (17):\penalty0 174102, 2002.

\bibitem[Hamilton et~al.(2017)Hamilton, Ying, and Leskovec]{hamilton2017inductive}
Will Hamilton, Zhitao Ying, and Jure Leskovec.
\newblock Inductive representation learning on large graphs.
\newblock \emph{Advances in neural information processing systems}, 30, 2017.

\bibitem[Behrendt et~al.(2019)Behrendt, Dimpfl, Peter, and Zimmermann]{behrendt2019rtransferentropy}
Simon Behrendt, Thomas Dimpfl, Franziska~J Peter, and David~J Zimmermann.
\newblock Rtransferentropy—quantifying information flow between different time series using effective transfer entropy.
\newblock \emph{SoftwareX}, 10:\penalty0 100265, 2019.

\bibitem[Sakoe and Chiba(1978)]{sakoe1978dynamic}
Hiroaki Sakoe and Seibi Chiba.
\newblock Dynamic programming algorithm optimization for spoken word recognition.
\newblock \emph{IEEE transactions on acoustics, speech, and signal processing}, 26\penalty0 (1):\penalty0 43--49, 1978.

\bibitem[Akturk et~al.(2021)Akturk, Dowd, Shankar, and Derdzinski]{akturk2021real}
Halis~Kaan Akturk, Robert Dowd, Kaushik Shankar, and Mark Derdzinski.
\newblock Real-world evidence and glycemic improvement using {D}excom {G6} features.
\newblock \emph{Diabetes Technology \& Therapeutics}, 23\penalty0 (S1):\penalty0 S--21, 2021.

\bibitem[Samsung R\&D Institute~Brazil et~al.(2023)Samsung R\&D Institute~Brazil, Penatti, Caetano, Cene, Sanches, Gomes, Cabrera, Beltrame, Lee, and Baek]{dvn_2023}
SRBR Samsung R\&D Institute~Brazil, Otávio A.~B. Penatti, Carlos Caetano, Vinicius~H. Cene, Ivandro Sanches, Victor~V. Gomes, Lizeth S.~B. Cabrera, Thomas Beltrame, Wonkyu Lee, and Sanghyun Baek.
\newblock {MIMIC-BP}.
\newblock \emph{Harvard Dataverse}, 2023.
\newblock \doi{10.7910/DVN/DBM1NF}.
\newblock URL \url{https://doi.org/10.7910/DVN/DBM1NF}.

\bibitem[Hubert and Levin(1976)]{hubert1976general}
Lawrence~J Hubert and Joel~R Levin.
\newblock A general statistical framework for assessing categorical clustering in free recall.
\newblock \emph{Psychological bulletin}, 83\penalty0 (6):\penalty0 1072, 1976.

\bibitem[Macqueen(1967)]{macqueen1967some}
J~Macqueen.
\newblock Some methods for classification and analysis of multivariate observations.
\newblock In \emph{Proceedings of 5-th Berkeley Symposium on Mathematical Statistics and Probability/University of California Press}, 1967.

\bibitem[Rousseeuw(1987)]{rousseeuw1987silhouettes}
Peter~J Rousseeuw.
\newblock Silhouettes: a graphical aid to the interpretation and validation of cluster analysis.
\newblock \emph{Journal of computational and applied mathematics}, 20:\penalty0 53--65, 1987.

\bibitem[Cali{\'n}ski and Harabasz(1974)]{calinski1974dendrite}
Tadeusz Cali{\'n}ski and Jerzy Harabasz.
\newblock A dendrite method for cluster analysis.
\newblock \emph{Communications in Statistics-theory and Methods}, 3\penalty0 (1):\penalty0 1--27, 1974.

\end{thebibliography}
